\documentclass[a4paper, 11pt]{article}

\usepackage[colorlinks=true, allcolors=blue]{hyperref}

\usepackage{cite}
\usepackage{apacite}
\AtBeginDocument{%
   
}

\usepackage[toc,title,page]{appendix}

\usepackage[a4paper, bindingoffset=0.2in,%
            left=1.0in, right=1.0in, top=1.0in, bottom=0.9in,%
            footskip=.25in]{geometry}

\usepackage[utf8]{inputenc}

\usepackage{booktabs}

\usepackage{multicol}
\usepackage{multirow}

\usepackage{xcolor}

\usepackage{graphicx}

\usepackage{float}
\usepackage{lscape}
\usepackage{subfigure}
\usepackage[flushleft]{threeparttable}
\usepackage{comment}

\author{
  P Krishna Kumar\\
  \href{mailto:kkumar@iihs.ac.in}{kkumar@iihs.ac.in}
  \and
  Krishnachandran Balakrishnan\\
  \href{mailto:kbalakrishnan@iihs.co.in}{kbalakrishnan@iihs.co.in}
  \and
  Pratyush Tripathy\\
  \href{mailto:pratyush@iihs.ac.in}{pratyush@iihs.ac.in}\\
  \em{\small Geospatial Lab, Indian Institute for Human Settlements, Bengaluru, India. 560 080}
}

\title{A CNN based method for Sub-pixel Urban Land Cover Classification using Landsat-5 TM and Resourcesat-1 LISS-IV Imagery}
\date{}

\begin{document}

\maketitle 

\begin{abstract}
Time series data of urban land cover is of great utility in analyzing urban growth patterns, changes in distribution of impervious surface and vegetation and resulting impacts on urban micro climate. While Landsat data is ideal for such analysis due to the long time series of free imagery, traditional per-pixel hard classification fails to yield full potential of the Landsat data. This paper proposes a sub-pixel classification method that leverages the temporal overlap of Landsat-5 TM and Resourcesat-1 LISS-IV sensors. We train a convolutional neural network to predict fractional land cover maps from 30m Landsat-5 TM data. The reference land cover fractions are estimated from a hard-classified 5.8m LISS-IV image for Bengaluru from 2011. Further, we demonstrate the generalizability and superior performance of the proposed model using data for Mumbai from 2009 and comparing it to the results obtained using a Random Forest classifier. For both Bengaluru (2011) and Mumbai (2009) data, Mean Absolute Percentage Error of our CNN model is in the range of 7.2 to 11.3 for both built-up and vegetation fraction prediction at the 30m cell level. Unlike most recent studies where validation is conducted using data for a limited spatial extent, our model has been trained and validated using data for the complete spatial extent of two mega cities for two different time periods. Hence it can reliably generate 30m built-up and vegetation fraction maps from Landsat-5 TM time series data to analyze long term urban growth patterns.
\end{abstract}

\section{Introduction}
The process of urbanization brings about significant changes in landscape pattern and land cover of the area concerned \shortcite{seto2011meta,weng2007spatiotemporal,jenerette2010global}. The rate of urban growth and urban spatial configuration can be precisely determined from remote sensing data due to its ability to provide repetitive and consistent coverage of the earth \shortcite{congalton2014global,donnay2000remote}. Hence, time-series analysis using remote sensing data is invaluable in understanding changes in landscape pattern and land cover transitions in urban areas and the resulting impacts on neighborhoods, urban ecosystems, and urban micro climate \shortcite{sultana2020assessment,pasquarella2016imagery,ridd1995exploring,tripathy2019monitoring}.

\begin{figure*}[ht!]
\centering
\includegraphics[width=\textwidth]{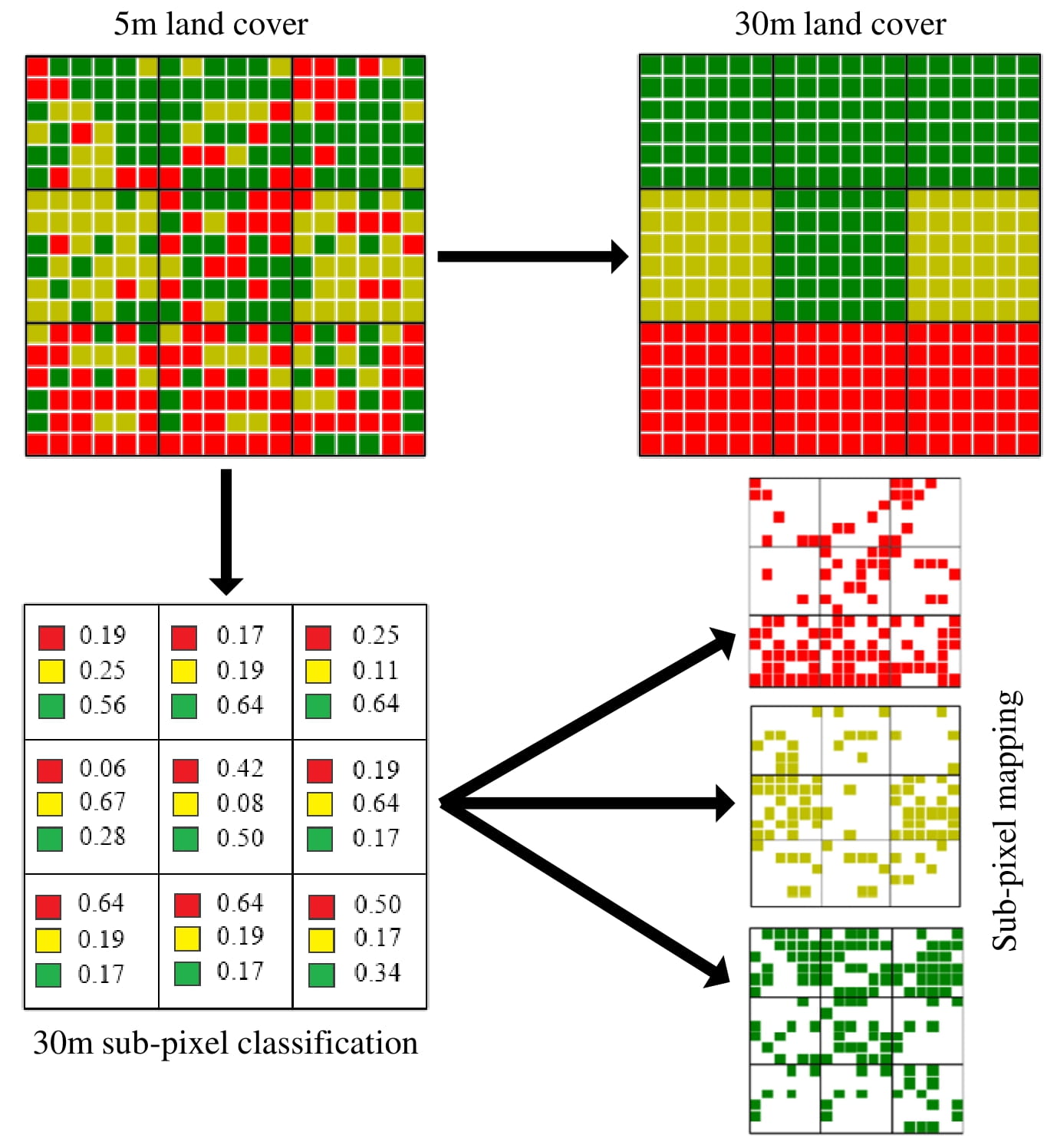}
\caption{Illustrating the difference between per-pixel and sub-pixel classification/mapping approach.}
\label{fig:fig1}
\end{figure*}

The Landsat sensors are ideal for such analysis due to the long time series of freely available data at medium resolution. The Landsat sensors have now been collecting data for almost 50 years and of these, Landsat-5 was the longest operating earth observation satellite and collected imagery from 1984 till 2011. However, they have limitation in pixel-based classification due to the presence of mixed pixels which occur due to the heterogeneity of land cover surface materials and surface structure which influence its reflectance \shortcite{atkinson1997mapping,maclachlan2017subpixel,kumar2010some}. To tackle this problem, the proportions of classes (sub-pixels) in these mixed pixels need to be estimated. This includes both sub-pixel classification and sub-pixel mapping or super-resolution methods \shortcite{genitha2019hybrid}. While the sub-pixel classification provides the relative abundance of land covers within a pixel as fractional values, the latter also provides the spatial distribution of these fractions within the pixel. The difference between per-pixel hard classification and sub-pixel approach is illustrated in Figure \ref{fig:fig1}. This paper discusses only sub-pixel classification and not sub-pixel mapping techniques.

Various approaches have been considered in the literature to address the mixed pixel problem. Spectral unmixing or endmember estimation methods are popular among them \shortcite{li2020mapping,powell2007sub,deng2013use,yuan2008comparison}. A linear mixture model is useful if the given mixed pixel appears in spatially segregated patterns of member classes \shortcite{weng2009estimating}. On the other hand, non-linear models are used if the member classes are tightly coupled \shortcite{somers2009nonlinear}. Multiple endmembers based and spatially adaptive unmixing techniques have proven to be effective in handling mixed pixels \shortcite{powell2007sub,deng2013spatially,kumar2017exploring}. However, spectral unmixing demands the generation of endmember spectra which is a difficult task particularly within heterogeneous urban environments. Moreover, the variations in spectral signatures of the underlying constituent materials over time can be caused by both acquisition (e.g. atmospheric, illumination) and seasonal changes \shortcite{zare2013endmember,somers2011endmember}. This puts a serious limitation particularly in time series analysis since too many endmembers may be required for adequately modelling this intra class spectral variability.

Both parametric and non-parametric regression models have been used to solve the mixed pixel problem \shortcite{okujeni2013support,walton2008subpixel,patidar2018multi,deng2017subpixel}. Non-parametric regression usually requires more training samples to supply both the model structure as well as the model estimates. Typically, machine learning methods are used for non-parametric non-linear regression due to its ability to extract the model of the mixture directly from the observed data \shortcite{bovolo2010novel,heremans2015machine,deng2017subpixel,hu2011estimating}. However, to obtain optimal performance from these methods, additional sample selection strategies are required besides the input data in the form of tabulated feature sets.

A convolutional neural network (CNN) differs from other machine learning methods that it can handle large amount of data and can capture multiple domain characteristics (spectral, temporal, resolution and directional) of images \shortcite{hu2018deep,petliak2019s}. Unlike standard linear regression techniques which reduce the bias at a cost of prediction accuracy, CNN explicitly address the bias-variance trade-off using a non-linear representational network \shortcite{mehta2019high}. Moreover, CNN automatically learns the best discriminatory features from images and hence avoiding their explicit measurement from the data. These potentials have motivated us for the consideration of a CNN model for sub-pixel classification of remote sensing imagery. In the literature, CNN has been used more for hyperspectral unmixing \shortcite{arun2018cnn,zhang2018hyperspectral}, as compared to sub-pixel classification of multispectral imagery.

\subsection{Contribution of this paper}
This study relies on the temporal overlap between the coarse resolution (30m) imagery from Landsat-5 TM with the 5.8m resolution (which is resampled to 5m) Linear Imaging Self Scanning sensor IV (LISS-IV) imagery from the Resourcesat-1 program by the Indian Space Research Organization (ISRO). We use this correspondence for reliable and accurate quantification of fractional built-up / impervious surface and vegetation in each (30mx30m) cell of the Landsat image. Since Landsat-5 TM was operational from 1985 to 2011 and Resourcesat-1 LISS-IV was operational from 2003 to 2013, it is highly feasible to do a direct mapping between them. For this, we have obtained cloud free Landsat-5 TM and LISS-IV images within a suitable temporal window. A major advantage of this strategy is that the same trained CNN model can be used to predict the fractional abundance from Landsat-5 TM image acquired at different time periods (1985-2011).

The paper is organized into six sections. The next section presents the details of study area and data preparation. Section \ref{sec3} details the sub-pixel classification process including the design specifics of CNN architecture. Section \ref{sec4} addresses the design of experiments for performance evaluation of the proposed technique. A detailed analysis of results and comparison with recent methods is presented in Section \ref{sec5}. Finally, concluding remarks are given in Section \ref{sec6}.

\section{Study Area and Data}\label{sec2}
\subsection{Study Area}\label{sec2.1}

\begin{figure*}[ht!]
\centering
\subfigure[]{\includegraphics[width=12cm]{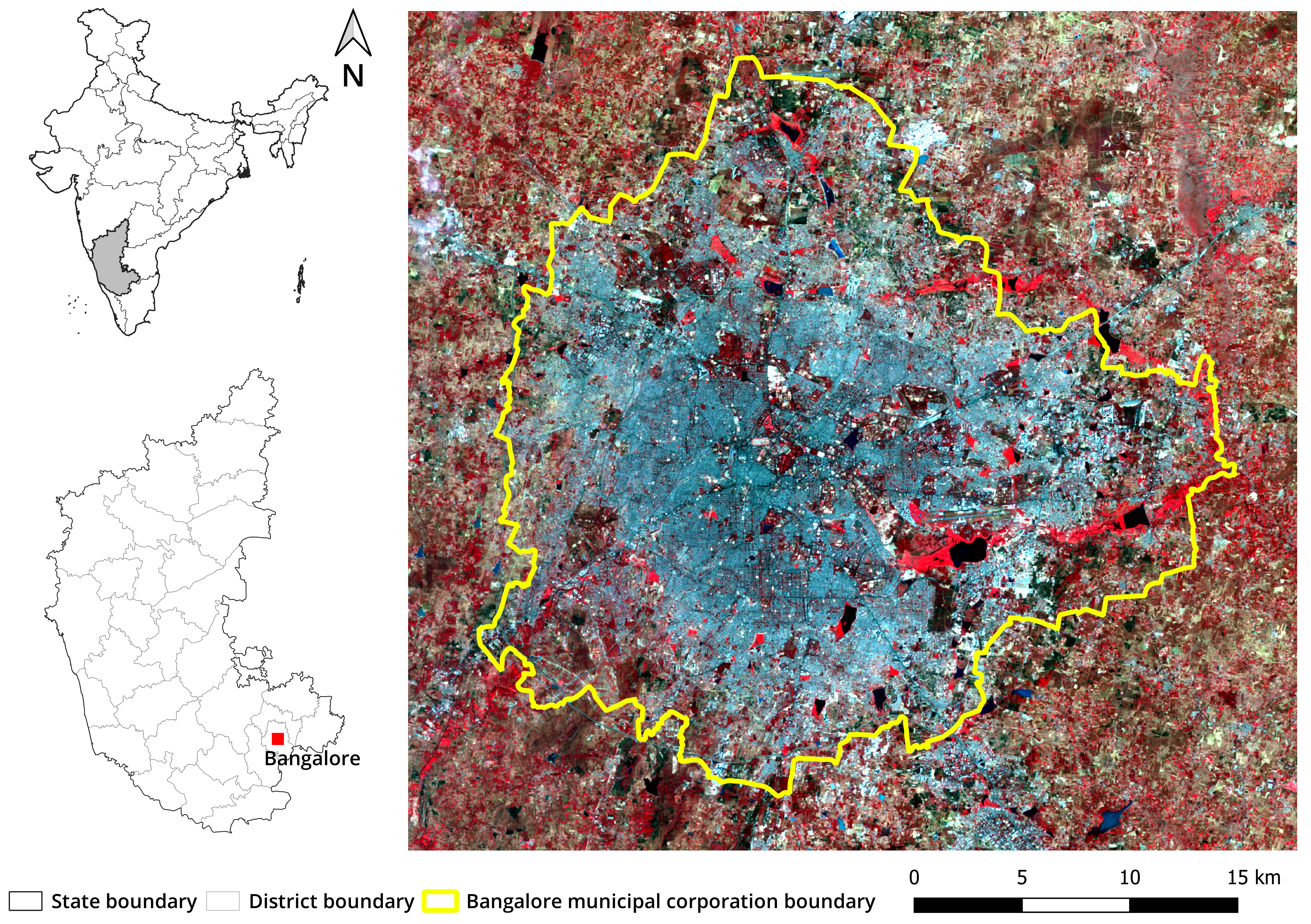}}
\subfigure[]{\includegraphics[width=12cm]{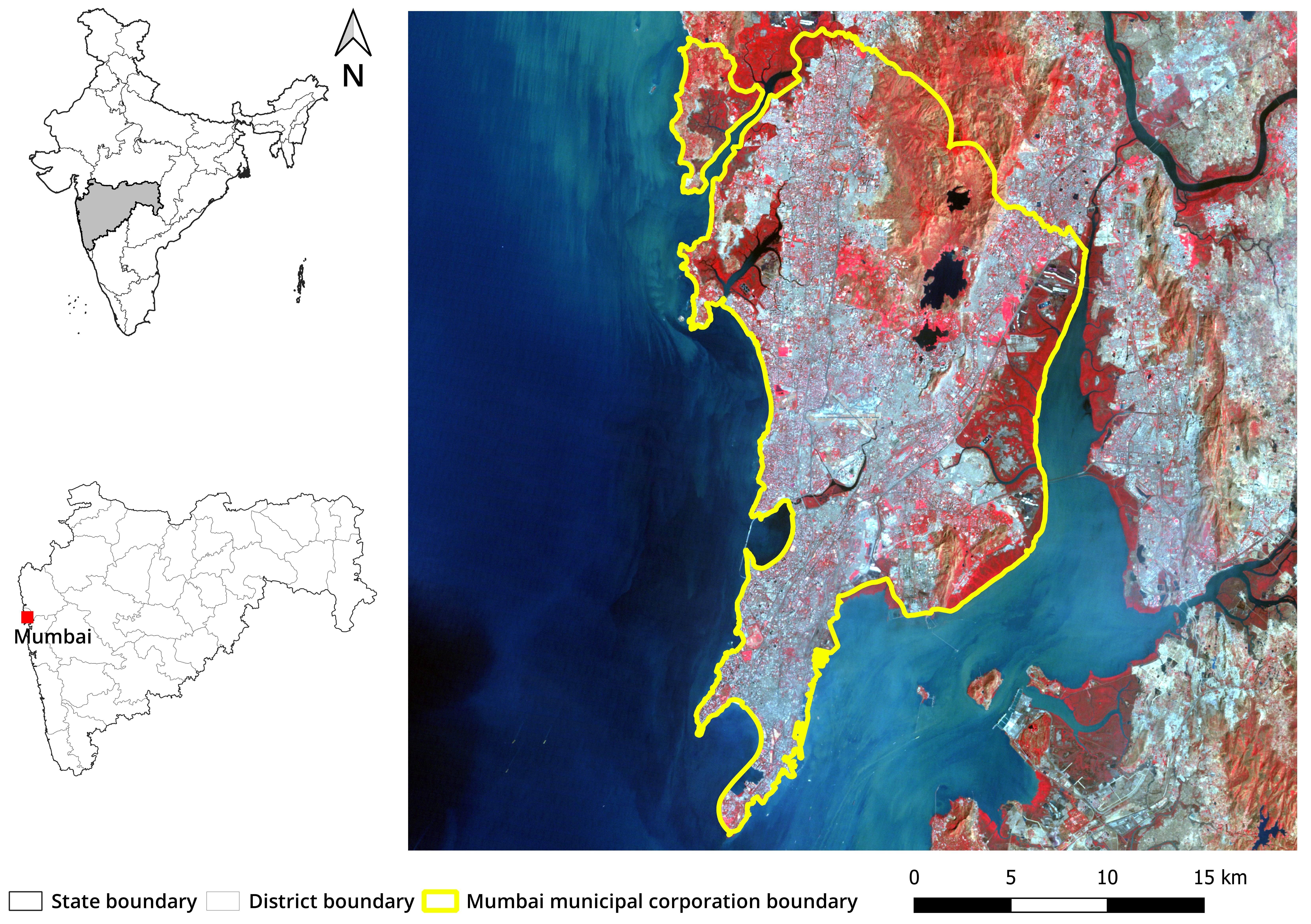}}
\caption{Study area: Landsat view of (a) Bengaluru and (b) Mumbai urban, India.}
\label{fig:fig2}
\end{figure*}

Satellite image data was collected for Bengaluru and Mumbai located in Southern and Western part of India. Bengaluru is the capital of the state of Karnataka, located between 77.45 and 77.79 decimal degrees longitudes, and 12.82 and 13.15 decimal degrees latitudes (see Figure \ref{fig:fig2}(a)). As one of the fastest growing cities in India, Bengaluru has undergone dramatic expansion in the past two decades. The city and its vicinity have a diverse range of land cover features comprising of various activities including but not limited to agriculture, stone quarrying and numerous lakes.  The city receives most of its rainfall in the June to October period, with average of 57 rainy days in a year. The average annual rainfall in Bengaluru as reported by the Indian Meteorological Department is 94cm \shortcite{imdbangalore}. The spectral characteristics of barren land during summer and dry season and stone quarries often match with that of the built impervious surface. Mumbai, located on the west coast of India, is the capital of the state of Maharashtra, lies between 72.77E and 72.98E decimal degrees longitudes, and 18.89N and 19.27N decimal degrees latitudes (see Figure \ref{fig:fig2}(b)). Constrained by water on three sides, the city continues to grow to the north-west direction. Sea beaches and large paved plots in the vicinity of dockyards on the eastern shoreline are the ones that are often falsely classified as built-up structures due to spectral inseparability. Much of the annual rainfall occurs during June through October, and the average annual rainfall is 205.3cm \shortcite{imdmumbai}. For both cities, vegetation, and inland water bodies are affected significantly due to the seasonal variation, which results in a change in their spectral characteristics in the satellite data.

\subsection{Data Preparation}
Landsat-5 TM imagery consists of seven spectral bands. The spatial resolution of bands 1 to 7 is 30 meters; whereas band-6 (thermal infrared) is captured at 120 meters, which is then resampled to 30 meters. We have tested our model with both Landsat Collection-1 Level-1 and Landsat Collection-1 Level-2 surface reflectance (atmospherically corrected) data. Since Collection 1 Level-2 data does not contain thermal band, we supplemented the 30m thermal band from level-1 data to keep the number of bands unchanged.

The LISS-IV sensor in Resourcesat-1 is a multi-spectral (MS) high-resolution sensor with a spatial resolution of 5.8 m at nadir. The data is acquired in three spectral bands namely visible (green and red) and near infrared covering a swath of 23.9 km in the MS mode. The National Remote Sensing Agency (NRSC), India, provides LISS-IV imagery after resampling to a cell size of 5m. Since the years of overlap between Landsat-5 and LISS-IV is from 2003 until 2011, we selected two pairs of images from years 2011 (for Bengaluru) and 2009 (for Mumbai). While the data from Bengaluru is used to train, test, and deploy the model, the latter is used to investigate the generalizability. Since, cloud free LISS-IV imagery was not available for the entire area covered by Landsat-5 TM, analysis was conducted after cropping Landsat-5TM to the area for which LISS-IV imagery were available.

The reference land cover maps at 5m were generated through unsupervised hard classification of the LISS-IV multi-spectral imagery using ISODATA clustering in QGIS 3.4.12 LTR. Further, we used Normalized Difference Vegetation Index (NDVI) to help in identification of vegetation class.  Since the reference land cover maps are of 5m resolution, we have 36 cells (6x6 grid) containing the true land cover information for each 30m cell of the corresponding Landsat-5 TM data. The fractional land cover data in each Landsat cell is computed directly using this overlap. The spatial shifts between the LISS-IV and Landsat-5 were corrected using the QGIS 3.4.12 LTR.

\begin{figure*}[ht!]
\centering
\subfigure[]{\includegraphics[width=11.3cm]{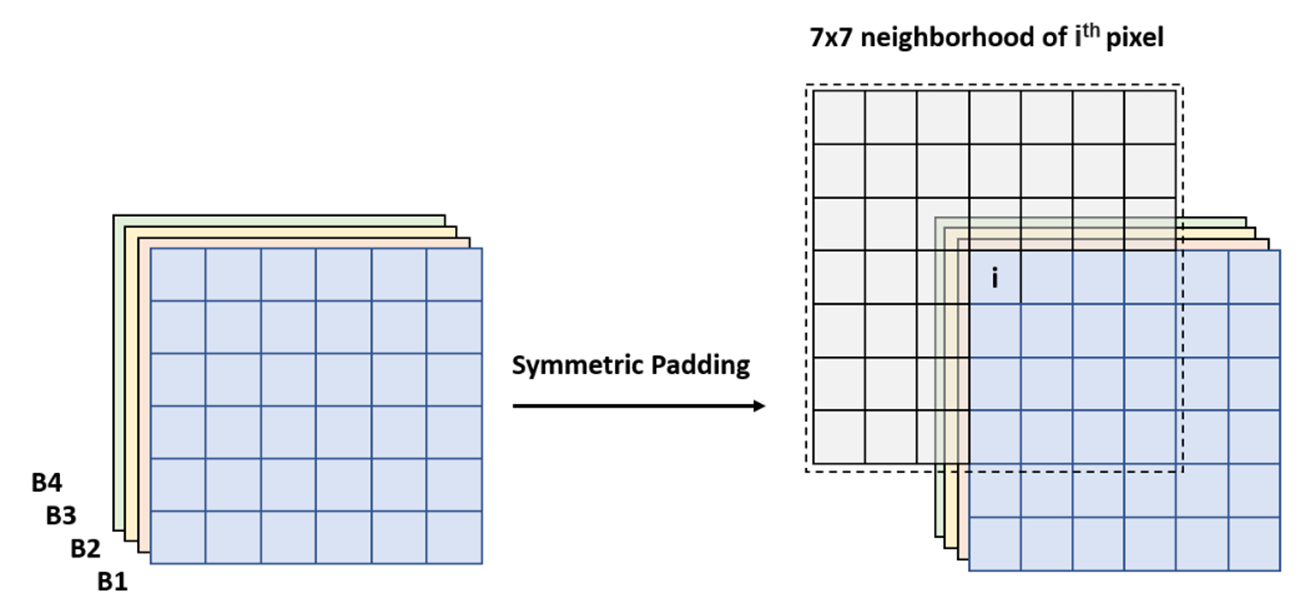}}
\subfigure[]{\includegraphics[width=11.3cm]{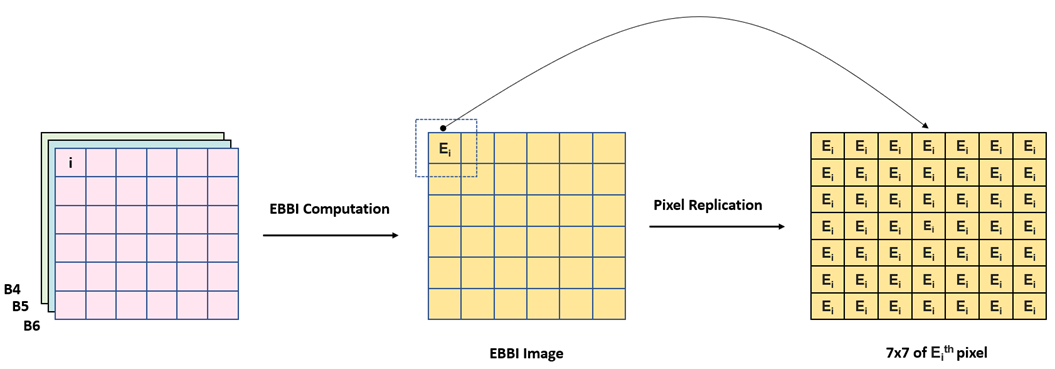}}
\subfigure[]{\includegraphics[width=11.3cm]{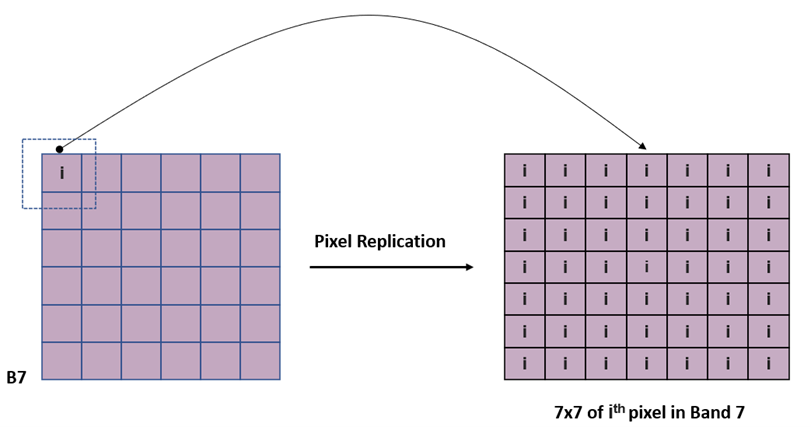}}
\caption{Input data preparation for the CNN model: 7x7 neighborhood of ith pixel after symmetric padding of band 1- band 4 (b) 7x7 of ith pixel in EBBI image and (c) 7x7 of ith pixel in band 7.}
\label{fig:fig3}
\end{figure*}

The CNN model that is designed for the sub-pixel classification can explore both spectral as well as spatial features of the training samples composed of image patches. Hence, we chose those spectral bands of Landsat-5 TM that sufficiently captures the spectral variability of land covers in the scene \shortcite{powell2007sub}. The first four bands (B1-B4) of image data are padded by replicating the border pixels (symmetric padding) and then divided it into 7x7 neighborhood of pixels (refer Figure \ref{fig:fig3}(a)). So, if there are ‘N’ number of Landsat pixels, then ‘N’ such neighborhoods will be found (7x7x4xN). This constitutes the first four inputs to the CNN. The Enhanced Built-Up and Bareness Index (EBBI) is computed as \shortcite{as2012enhanced}:

\begin{equation}
\label{eq:eq1}
EBBI = \frac{Band_5 - Band_4}{10\sqrt{Band_5+Band_6}} 
\end{equation}
The $EBBI$ value is calculated for each pixel without considering the neighborhood, and hence, no image padding is required. But the obtained EBBI value corresponding to each pixel is replicated to match with the pixel neighborhood size (7x7) as shown in Figure \ref{fig:fig3}(b). Since there are ‘N’ number of Landsat pixels, this forms (7x7x1xN) the fifth input to the CNN. In a similar way, band 7 (SWIR-2) pixels are also replicated and matched with the pixel neighborhood size (7x7x1xN) (refer Figure \ref{fig:fig3}(c)). Thus, the input to the CNN is a 4D data tensor (7×7×6×N), where the first and second dimensions correspond to the size of the spatial neighborhood of the image data and the third dimension corresponds to spectral feature dimension. The last dimension implies the number of input training samples (number of Landsat pixels).

By computing the EBBI ratio, we are merging the information from band 6 with the band 5 to reduce the number of inputs to the CNN. Further, the EBBI is a widely used remote sensing index which effectively discriminates built-up and bare land areas. This allows faster convergence of the CNN model. We have not considered the neighborhoods of pixels in band 5 (B5, SWIR-1), band 6 (B6, thermal band) and band 7 (B7, SWIR-2), as these are the most sensitive bands in separating the LULC class type \shortcite{biradar2007establishing}. Further, the thermal band is only available at a lower resolution (120m) which makes the use of its neighborhood level data less significant, but at the same time contain valuable information related to the spatial variations of land surface and vegetation properties \shortcite{defries2000global,foody1996approaches}.

\section{Materials and Methods}\label{sec3}
\subsection{Overview}\label{sec3.1}
\begin{figure*}[ht]
\centering
\includegraphics[width=14cm]{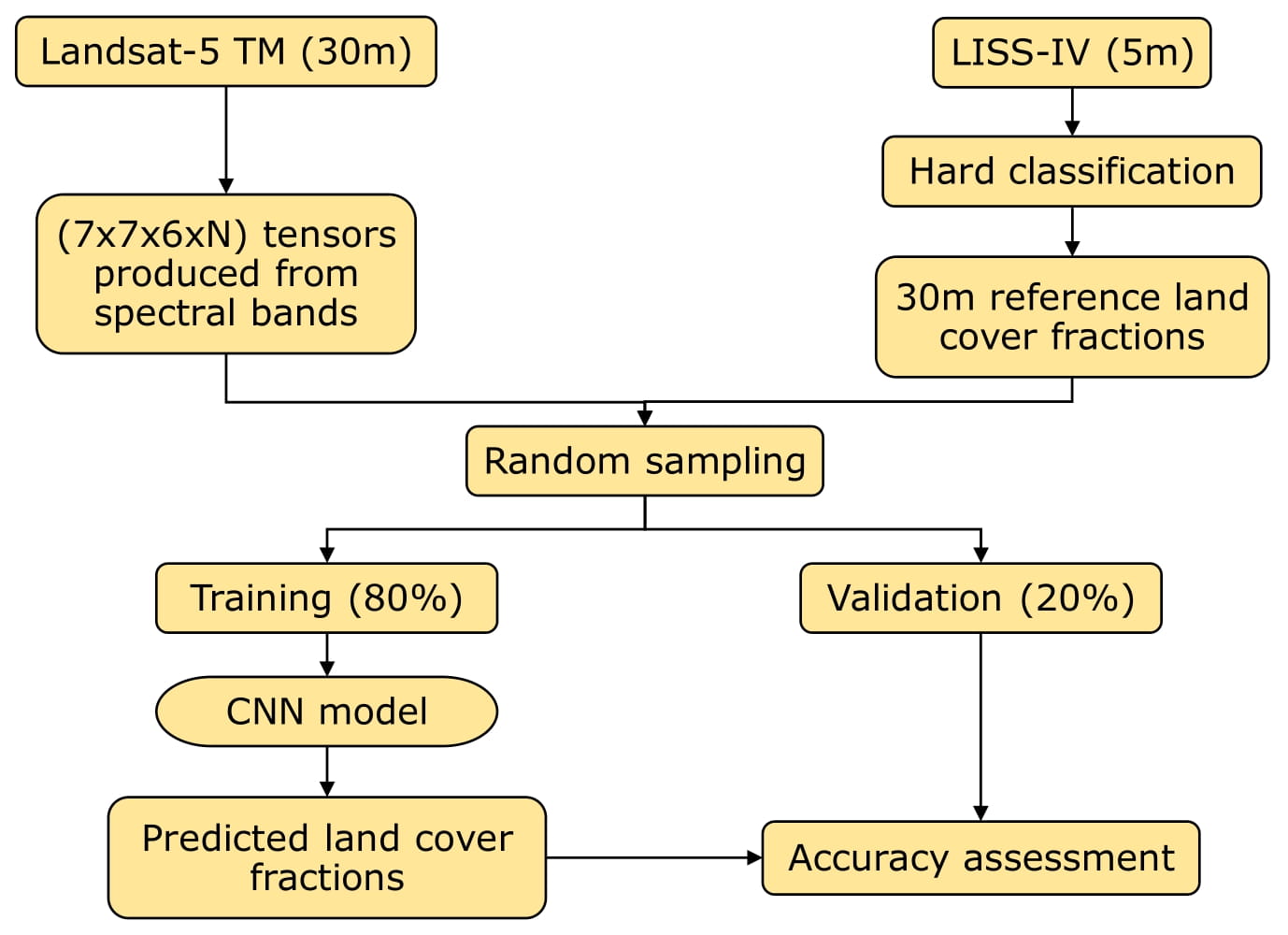}
\caption{Flow chart of the proposed method: Predictions of the land cover fractions using a CNN model.}
\label{fig:fig4}
\end{figure*}

Flowchart shown in Figure \ref{fig:fig4} gives an overview of the proposed method. Before we begin, both Landsat-5 TM and LISS-IV images are cropped to the same extent such that each 30mx30m Landsat cell is perfectly overlaid on the 6x6 grid (36 cells) of LISS-IV image. This arrangement facilitates the computation of land cover fractions in these Landsat cells (reference land cover fractions) which are estimated independently from each 6x6 grid of hard classified LISS-IV map. The spectral features are extracted from Landsat-5 TM in the form of a 4D data tensor (7x7x6xN) where N denotes the number of Landsat cells. The training and validation data are separated using a random sampling. A CNN model is then trained and assessed its accuracy by comparing the predictions with the reference land cover fractions.

\subsection{CNN Architecture}\label{sec3.2}
\begin{figure*}[ht]
\centering
\includegraphics[width=\textwidth]{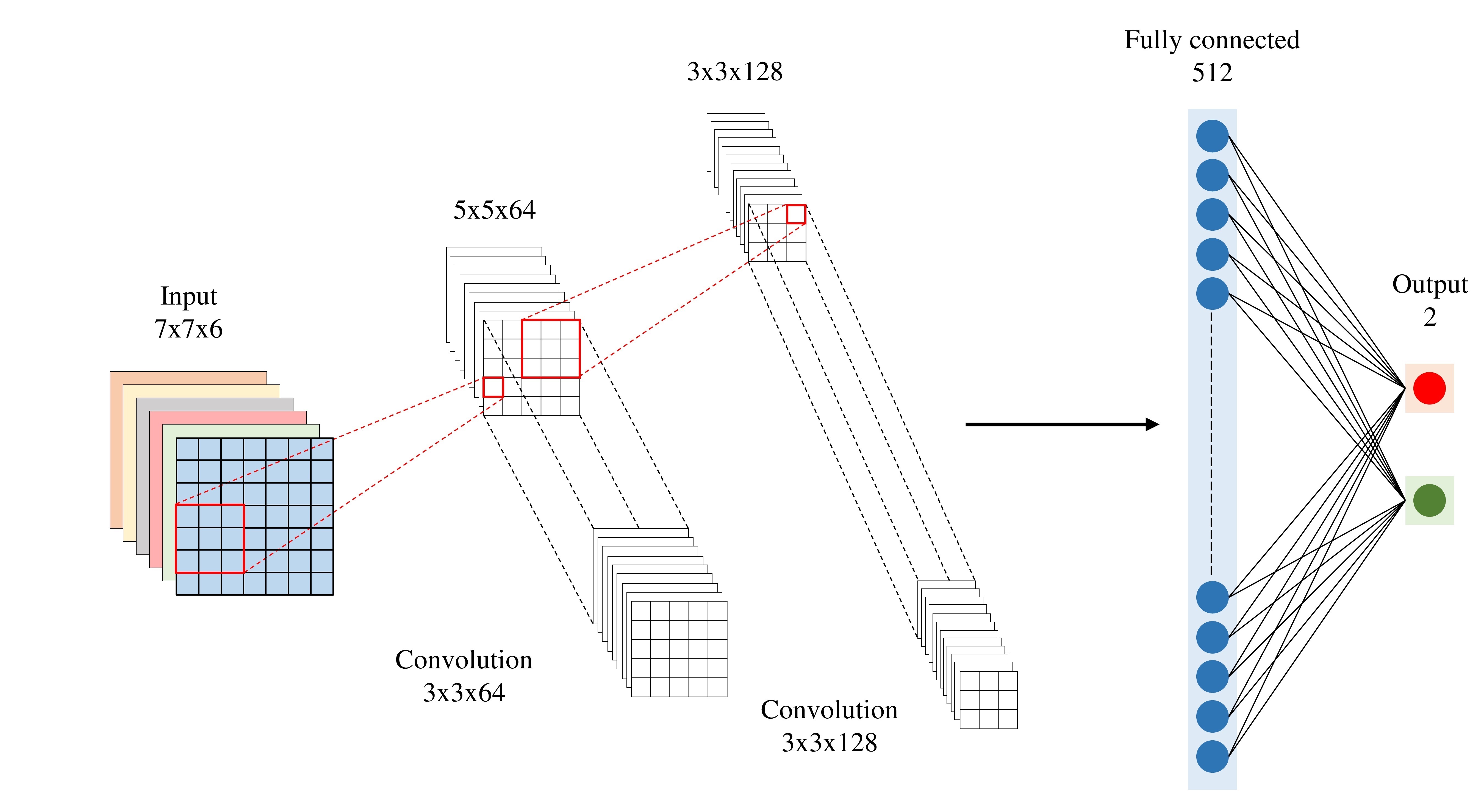}
\caption{CNN architecture for the sub-pixel classification.}
\label{fig:fig5}
\end{figure*}

We have used the Keras neural network library written in Python to build the CNN model. The network architecture is illustrated in Figure \ref{fig:fig5} which is composed of convolutional layers (CL1 and CL2), dropout layers, and fully connected layers (FC1 and FC2). The initial weights of the network are computed from a Gaussian distribution with a standard deviation of 0.05 and a mean value of 0. First two convolutional layers with 64 and 128 kernels help to extract useful features from input images. A kernel size of (3×3) pixel block is selected after several experiments. Since most of the useful features in an image are usually local, it makes sense to take few local pixels at a time to apply convolutions. Furthermore, increase in kernel size increases the number of learnable parameters and hence it is common for the convolutional kernel to be of size 3x3. Batch normalization is applied after CL1 and CL2 to reduce the internal covariate shift and speed up the training process. Further, dropout technique is applied to the CL1, CL2 and FC1 layer with probability values 0.5, 0.25 and 0.5, respectively to control over-fitting. Convolutional layers are followed by two fully connected layers each of which has 512 and 2 outputs. The final two outputs represent built-up and vegetation fraction of the respective Landsat cell.

‘Leaky ReLU’ activation function is used for both convolutional layers and fully connected layers to eliminate the “dying ReLU” problem. However, the output range of ‘Leaky ReLU’ is from -infinity to infinity, and hence, the final output value can go less than 0 or greater than 1. This makes it necessary to clip the output in the range [0, 1] to ensure that the fractions of the two classes are between 0 and 1. Further, no constraints were applied to restrict the fractional sum of land cover classes to unity. There are two reasons for this. In the first place, there may exist more land cover classes in the study area than built-up and vegetation quantifying which is not the scope of this research. Secondly, since the remote sensed spectral information in one 30mX30m Landsat cell is derived from complex scattering effects from background and foreground layers, like in the case of tree canopy over an impervious paved area, according to \shortcite{walton2008subpixel}, it is desirable to allow the cover proportions to add to more than 100 percent.

\subsection{CNN Training}\label{sec3.3}

A Dell laptop with CPU of Intel 8650U (4 Cores, 4.2 GHz), memory of 16 GB and graphics card of NVIDIA GeForce MX130 (4 GB) has been used for calculation. The training, testing and validation samples were selected via random sampling. The divisions were: 80\% for training, 5\% for validation and the remaining 15\% for testing. We considered a total of 15000 random samples from the study area, out of which 12000 samples were used for training, 750 for validation and 2250 for testing. In the training process, we set the batch size as 256 and the number of training epochs to 250 which we found gave the best results. We have not applied any early stopping criteria as it sometimes results in stopping the training process with a large error value. The CNN framework achieved maximum accuracy when Adam optimizer is used along with ‘logcosh’ loss function.

\subsection{Methods for Accuracy Assessment}\label{sec3.4}
Given that many mixed pixels are located in between classes, the accuracy assessment at sub-pixel level needs more careful selection of performance metrics. This includes visual as well as quantitative assessment techniques such as computation of mean absolute error (MAE), root mean square error (RMSE) and Nash–Sutcliffe (NS) index, box plots and bias histograms. In addition, we have used receiver operating characteristic (ROC) plots to assess the diagnostic ability of the prediction system for a binary classification at 50\% threshold level. Visual interpretation of land cover data is primarily for the purpose of map validation at different scales. The analysis of fractional land cover maps reveals the distribution of classes within each pixel. Further, this enables the analysis of the spatial structure of the image in a fine scale manner which provides a more realistic and meaningful representation of the landscape. We implemented 3-fold cross validation to evaluate our model while computing the percentage MAE, percentage RMSE, and NS index using below equations \shortcite{nash1970river}:

\begin{equation}
\label{eq:eq2}
MAE (\%) = \sum_{i=1}^{N} \frac{|Z_R-Z_P|}{N}*100 
\end{equation}

\begin{equation}
\label{eq:eq3}
RMSE (\%) = \left[ \sum_{i=1}^{N} \frac{(Z_R-Z_P)^2}{N} \right]^{\frac{1}{2}} *100 
\end{equation}
 
\begin{equation}
\label{eq:eq4}
NS = 1 - \left( \frac{\sum_{i=1}^{N}(Z_R - Z_P)^2}{\sum_{i=1}^{N}(Z_R - \overline Z)^2} \right)
\end{equation} 
                                                                   
where $Z_R$ represents the reference fraction, $Z_P$ represents the predicted fraction, $\overline Z$ is the average of the reference fractions and $N$ is the total number of samples used. The NS index is a normalized statistic that determines the relative magnitude of the residual variance (noise) compared to the measured data variance (information) \shortcite{nash1970river}. NS index is computed in addition to the coefficient of determination (R2) values, since R2 does not consider the fact that perfect prediction assumes an intercept equal to zero and a slope equal to one, which the NS index do \shortcite{heremans2015machine}. An NS index value of 1 indicates a perfect match between the estimated and the reference fractions.

In addition to conducting accuracy assessment at the original pixel size (30m x 30m), as discussed in \shortciteA{shih2020estimating,patidar2018multi,patidar2020rule,deng2017subpixel,deng2019examining,maclachlan2017subpixel,wu2014spatially,hu2011estimating}, we perform accuracy assessment on 90m × 90m (3 × 3 Landsat pixels) sample units. This technique of spatially averaging over different size spatial windows helps to decrease influence of registration error between high-resolution image and the multi-spectral images.

\section{Results}\label{sec4}
\subsection{Accuracy Assessment of the Reference Data}\label{sec4.1}
For accuracy assessment, we sampled 400 random cells within the image extent, separately for each image, and labelled the ground truth values by visual interpretation of the LISS-IV data. The accuracy values and respective confusion matrices were given in Table \ref{table1} along with Cohen’s kappa coefficients \shortcite{cohen1960coefficient}.

\begin{table}[htb]
\label{table1}
\caption{Confusion matrices for the reference land cover classifications.}
\vskip0.2cm
\begin{tabular}{|c|c|ccccc|}
\hline
\multirow{8}{*}{\begin{tabular}[c]{@{}c@{}}Bengaluru\\    \\ (2011)\end{tabular}} & \multirow{8}{*}{Predicted} & \multicolumn{5}{c|}{Actual}                                                                                                                          \\ \cline{3-7} 
                                                                                  &                            & \multicolumn{1}{c|}{Landcover}           & \multicolumn{1}{c|}{Built-up} & \multicolumn{1}{c|}{Vegetation} & \multicolumn{1}{c|}{Others} & Row total \\ \cline{3-7} 
                                                                                  &                            & \multicolumn{1}{c|}{Built-up}            & \multicolumn{1}{c|}{132}      & \multicolumn{1}{c|}{5}          & \multicolumn{1}{c|}{34}     & 171       \\ \cline{3-7} 
                                                                                  &                            & \multicolumn{1}{c|}{Vegetation}          & \multicolumn{1}{c|}{2}        & \multicolumn{1}{c|}{105}        & \multicolumn{1}{c|}{14}     & 121       \\ \cline{3-7} 
                                                                                  &                            & \multicolumn{1}{c|}{Others}              & \multicolumn{1}{c|}{7}        & \multicolumn{1}{c|}{1}          & \multicolumn{1}{c|}{100}    & 108       \\ \cline{3-7} 
                                                                                  &                            & \multicolumn{1}{c|}{Column total}        & \multicolumn{1}{c|}{141}      & \multicolumn{1}{c|}{111}        & \multicolumn{1}{c|}{148}    & 400       \\ \cline{3-7} 
                                                                                  &                            & \multicolumn{1}{c|}{Accuracy (\%)}       & \multicolumn{1}{c|}{93.6}     & \multicolumn{1}{c|}{94.6}       & \multicolumn{1}{c|}{67.6}   &           \\ \cline{3-7} 
                                                                                  &                            & \multicolumn{1}{c|}{Kappa   Coefficient} & \multicolumn{4}{c|}{0.76}                                                                                 \\ \hline
\multirow{6}{*}{\begin{tabular}[c]{@{}c@{}}Mumbai\\    \\ (2009)\end{tabular}}    & \multirow{6}{*}{Predicted} & \multicolumn{1}{c|}{Built-up}            & \multicolumn{1}{c|}{143}      & \multicolumn{1}{c|}{1}          & \multicolumn{1}{c|}{30}     & 174       \\ \cline{3-7} 
                                                                                  &                            & \multicolumn{1}{c|}{Vegetation}          & \multicolumn{1}{c|}{3}        & \multicolumn{1}{c|}{104}        & \multicolumn{1}{c|}{19}     & 126       \\ \cline{3-7} 
                                                                                  &                            & \multicolumn{1}{c|}{Others}              & \multicolumn{1}{c|}{13}       & \multicolumn{1}{c|}{5}          & \multicolumn{1}{c|}{82}     & 100       \\ \cline{3-7} 
                                                                                  &                            & \multicolumn{1}{c|}{Column total}        & \multicolumn{1}{c|}{159}      & \multicolumn{1}{c|}{110}        & \multicolumn{1}{c|}{131}    & 400       \\ \cline{3-7} 
                                                                                  &                            & \multicolumn{1}{c|}{Accuracy (\%)}       & \multicolumn{1}{c|}{89.9}     & \multicolumn{1}{c|}{94.5}       & \multicolumn{1}{c|}{62.6}   &           \\ \cline{3-7} 
                                                                                  &                            & \multicolumn{1}{c|}{Kappa   Coefficient} & \multicolumn{4}{c|}{0.73}                                                                                 \\ \hline
\end{tabular}
\end{table}

\subsection{Visual Assessment of the Landcover Fractions}\label{sec4.2}
\begin{figure}[ht!]
  \centering
  \includegraphics[width=15cm]{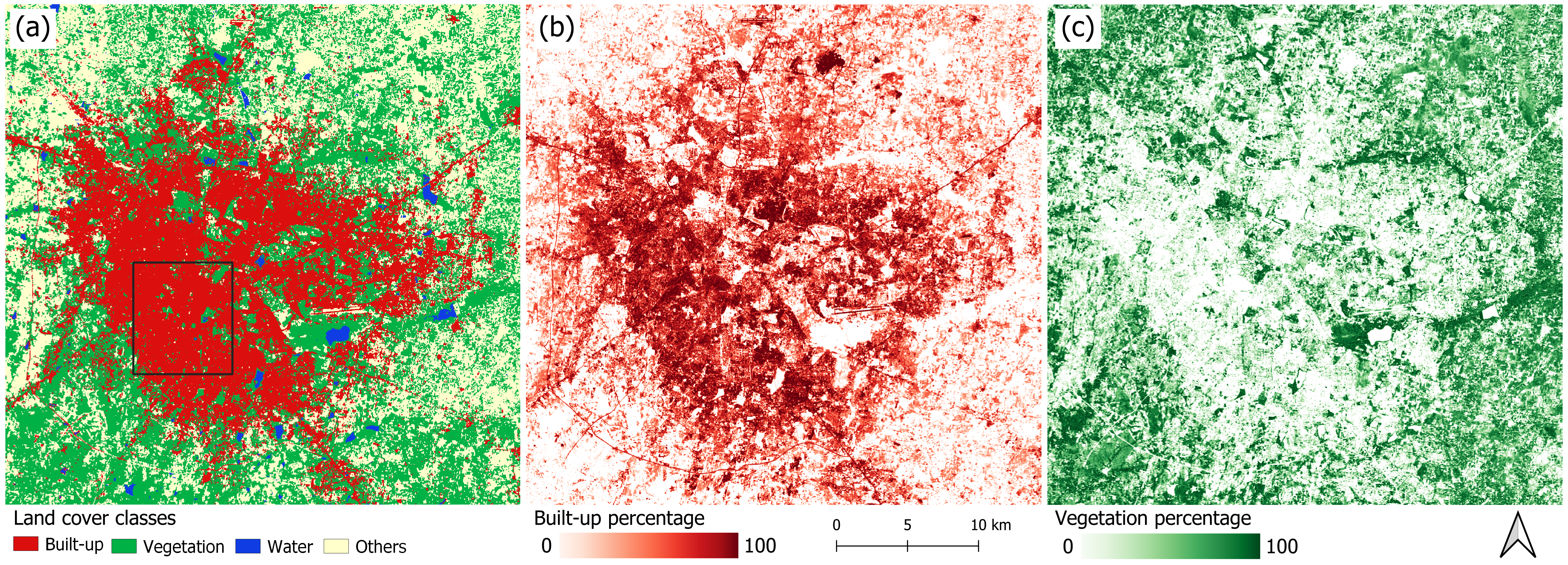}
  \caption{(a) 2011 land cover map for Bengaluru area at 30m resolution; (b) predicted percentage built-up and (c) predicted percentage vegetation for the same area.}
  \label{fig:fig6}

  \vspace*{1cm}

  \includegraphics[width=15cm]{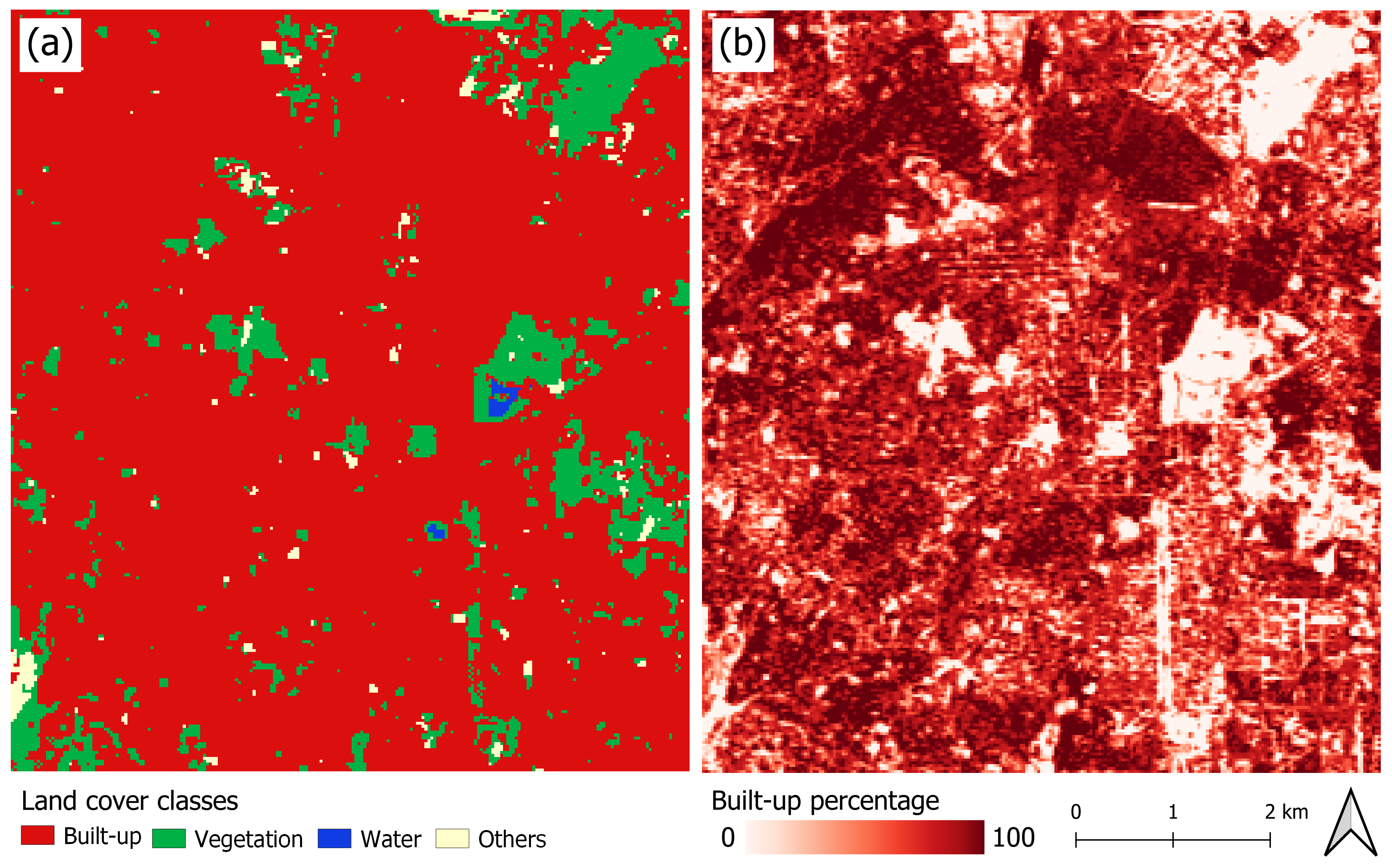}
  \caption{(a) Enlarged view of area within the black rectangle in Figure \ref{fig:fig6}(a); (b) enlarged view of the corresponding area from Figure \ref{fig:fig6}(b).}
  \label{fig:fig7}
  
\end{figure}

\begin{figure*}[ht]
\centering
\includegraphics[width=15cm]{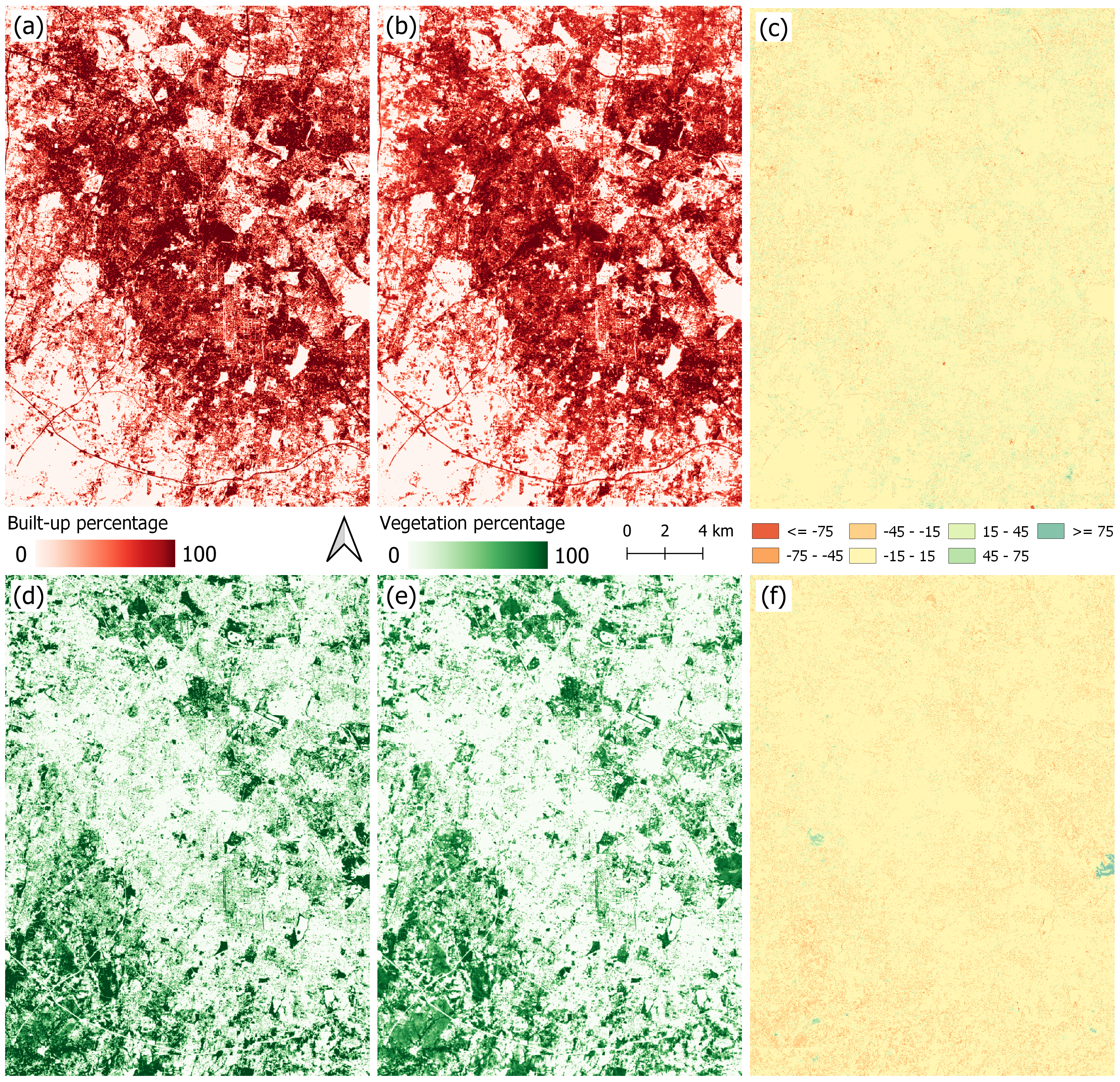}
\caption{30m resolution fractional maps of built-up and vegetation for Bengaluru (2011); (a) reference built-up fraction map; (b) predicted built-up fraction map; (c) prediction error map for built-up; (d) reference vegetation fraction map; (e) predicted vegetation fraction map; (f) prediction error map for vegetation.}
\label{fig:fig8}
\end{figure*}

Figure \ref{fig:fig6}(a) shows the 30m land cover data for Bengaluru derived from Landsat-5 TM for the year 2011 while Figure \ref{fig:fig6}(b) and Figure \ref{fig:fig6}(c) shows the predicted percentage of built-up and vegetation for each of the 30m Landsat cells, respectively. It can be observed that the 30m land cover data is not completely successful in demonstrating the heterogeneity and land cover mix in the city. This is clearer from Figure \ref{fig:fig7}(a) and Figure \ref{fig:fig7}(b) which shows an enlarged view of a small part from Figure \ref{fig:fig6}(a) and Figure \ref{fig:fig6}(b), respectively. From Figure \ref{fig:fig7}(b), we can identify various neighborhood types based on the percentage of built-up in each 30m cell. The fractional maps of built-up and vegetation in Fig. 8 illustrates the effectiveness of the proposed method. Figure \ref{fig:fig8}(a) shows the reference built-up fraction derived from the hard classified LISS-IV image and Figure \ref{fig:fig8}(b) shows the predicted built-up fraction. Figure \ref{fig:fig8}(d) and Figure \ref{fig:fig8}(e) shows the vegetation counterparts. Respective reference and predicted fraction maps are quite similar except at a few locations which can be easily recognized from the spatial error distribution maps shown in Figure \ref{fig:fig8}(c) and Figure \ref{fig:fig8}(f) which are produced by subtracting the reference fraction from the predicted fraction.

\subsection{Quantitative Analysis}\label{sec4.3}
The performance of the model was evaluated on both Landsat Collection-1 Level-1 and Landsat Collection-1 Level-2 surface reflectance data. This is to avoid any possible errors due to the atmospheric effects in the Level-1 data. However, we have not observed any significant changes in the performance. Hence, we present the results using the Level-1 data. The same argument is applicable to the results presented in Section \ref{sec4.4} and Section \ref{sec4.5}.

\subsubsection{System performance}
With the 15000 random samples that we considered, the CNN took nearly 215 seconds for training (12000 Landsat pixels) and less than 3 seconds for prediction (2250 Landsat pixels). In Table \ref{table2}, performance of the CNN model is compared against a model without any hidden layer (only input and output layer). This model without hidden layer acts like a linear regression (LR) model. From Table \ref{table2}, we can conclude that the CNN model is a substantial improvement over the LR model in estimating the built-up and vegetation fractions.

\subsubsection{3-fold cross validation}
We implemented a 3-fold cross validation system to evaluate the generalizability of our model architecture. We split our 15,000 random samples into three different subgroups. Each of Fold-1 and Fold-2 had 35\% of samples (5250 samples each), while Fold-3 had 30\% of data samples (4500 samples). Then three models with the same architecture as before were trained and evaluated with each fold given a chance to be held-out as test set as described below:
\begin{itemize}
    \item Model-1: Trained on Fold-1 $+$ Fold-2, Tested on Fold-3
    \item Model-2: Trained on Fold-2 $+$ Fold-3, Tested on Fold-1
    \item Model-3: Trained on Fold-1 $+$ Fold-3, Tested on Fold-2
\end{itemize}

As evident from Table \ref{table3}, the cross validation provides consistent results which are comparable to the results of the original model given in Table \ref{table2}. The slightly lower performance in cross validation can be attributed to training with fewer samples.

\begin{table}[ht]
\caption{Performance comparison of the CNN model with LR model.}
\label{table2}
\begin{tabular}{|c|c|c|c|c|c|c|}
\hline
                           & Pixel size                                                                                  & Landcover type & MAE (\%) & RMSE (\%) & R2    & NSI  \\ \hline
\multirow{4}{*}{CNN model} & \multirow{2}{*}{\begin{tabular}[c]{@{}c@{}}30 m × 30 m\\  (single pixel)\end{tabular}} & Built-up       & 9.1      & 13.9      & 0.88* & 0.88 \\ \cline{3-7} 
                           &                                                                                             & Vegetation     & 7.2      & 12.8      & 0.88* & 0.87 \\ \cline{2-7} 
                           & \multirow{2}{*}{\begin{tabular}[c]{@{}c@{}}90 m × 90 m\\  (3×3 pixels)\end{tabular}}   & Built-up       & 6.0      & 10.5      & 0.92* & 0.92 \\ \cline{3-7} 
                           &                                                                                             & Vegetation     & 4.5      & 7.2       & 0.94* & 0.94 \\ \hline
\multirow{2}{*}{LR model}  & \multirow{2}{*}{\begin{tabular}[c]{@{}c@{}}30 m × 30 m\\ (single pixel)\end{tabular}} & Built-up       & 15.3     & 24.5      & 0.75* & 0.76 \\ \cline{3-7} 
                           &                                                                                             & Vegetation     & 14.5     & 21.3      & 0.76* & 0.76 \\ \hline
\end{tabular}

\begin{tablenotes}
\centering
      \small
      \item *Statistically significant relationship $(p < 0.01)$
    \end{tablenotes}

\end{table}


\begin{table}[]
\centering
\caption{Results of 3-fold cross-validation.}
\label{table3}
\vskip0.1cm
\begin{tabular}{|l|l|cc|cc|}
\hline
\multicolumn{1}{|c|}{\multirow{2}{*}{Pixel   size}}                                           & \multicolumn{1}{c|}{\multirow{2}{*}{Model}} & \multicolumn{2}{c|}{MAE (\%)}              & \multicolumn{2}{c|}{RMSE (\%)}             \\ \cline{3-6} 
\multicolumn{1}{|c|}{}                                                                        & \multicolumn{1}{c|}{}                       & \multicolumn{1}{c|}{Built-up} & Vegetation & \multicolumn{1}{c|}{Built-up} & Vegetation \\ \hline
\multirow{3}{*}{\begin{tabular}[c]{@{}l@{}}30   m × 30 m\\  (single pixel)\end{tabular}} & Model-1                                     & \multicolumn{1}{c|}{9.1}      & 8.2        & \multicolumn{1}{c|}{14.8}     & 13.2       \\ \cline{2-6} 
                                                                                              & Model-2                                     & \multicolumn{1}{c|}{9.0}      & 7.7        & \multicolumn{1}{c|}{14.8}     & 12.9       \\ \cline{2-6} 
                                                                                              & Model-3                                     & \multicolumn{1}{c|}{8.8}      & 7.8        & \multicolumn{1}{c|}{14.6}     & 13.0       \\ \hline
\end{tabular}
\end{table}

\begin{table}[ht!]
\caption{System performance with different neighborhood window sizes.}
\label{table4}
\resizebox{\textwidth}{!}{%
\begin{tabular}{|l|c|c|cc|cc|}
\hline
\multicolumn{1}{|c|}{\multirow{2}{*}{Pixel   size}}                                         & \multicolumn{1}{l|}{\multirow{2}{*}{\begin{tabular}[c]{@{}l@{}}Neighborhood\\ Window Size\end{tabular}}} & \multirow{2}{*}{\begin{tabular}[c]{@{}c@{}}Filter Kernel sizes\\ (CL1, CL2)\end{tabular}} & \multicolumn{2}{c|}{MAE (\%)}              & \multicolumn{2}{c|}{RMSE (\%)}             \\ \cline{4-7} 
\multicolumn{1}{|c|}{}                                                                      & \multicolumn{1}{l|}{}                                                                                          &                                                                                                 & \multicolumn{1}{c|}{Built-up} & Vegetation & \multicolumn{1}{c|}{Built-up} & Vegetation \\ \hline
\multirow{4}{*}{\begin{tabular}[c]{@{}l@{}}30 m × 30 m\\ (single pixel)\end{tabular}} & 3x3                                                                                                            & (3x3,1x1)                                                                                       & \multicolumn{1}{c|}{19.3}     & 9.4        & \multicolumn{1}{c|}{25.7}     & 13.6       \\ \cline{2-7} 
                                                                                            & 5x5                                                                                                            & (3x3,3x3)                                                                                       & \multicolumn{1}{c|}{11.7}     & 8.0        & \multicolumn{1}{c|}{18.7}     & 13.2       \\ \cline{2-7} 
                                                                                            & 7x7                                                                                                            & (3x3,3x3)                                                                                       & \multicolumn{1}{c|}{9.1}      & 7.2        & \multicolumn{1}{c|}{13.9}     & 12.8       \\ \cline{2-7} 
                                                                                            & 9x9                                                                                                            & (5x5,3x3)                                                                                       & \multicolumn{1}{c|}{11.4}     & 8.6        & \multicolumn{1}{c|}{18.2}     & 14.1       \\ \hline
\end{tabular}%
}
\end{table}

\subsubsection{Performance evaluation with varying neighborhood window sizes}
To decide the optimum neighborhood window size, we tested our model with different window sizes such as 3x3, 5x5, 7x7 and 9x9. While doing this, we modified the kernel sizes in our CNN model to match with the neighborhood window size. We observed that a window size of 3x3 results in an underestimation of the built-up and vegetation fractions whereas 9x9 results in an overestimation. But concurrently, as the window size increased, the misclassifications got reduced significantly which improved the accuracy values. Here, what we mean by underestimation is that majority of the predicted values are lesser than the reference fraction values. Similarly, overestimation means majority of the predicted values are greater than the reference fraction values. A neighborhood window size of 7x7 delivered a balance between the two which reflected in the results in Table \ref{table4}.

\subsubsection{Boxplots and bias histograms}
\begin{figure*}[t!]
\centering
\subfigure[]{\includegraphics[width=1.0\textwidth]{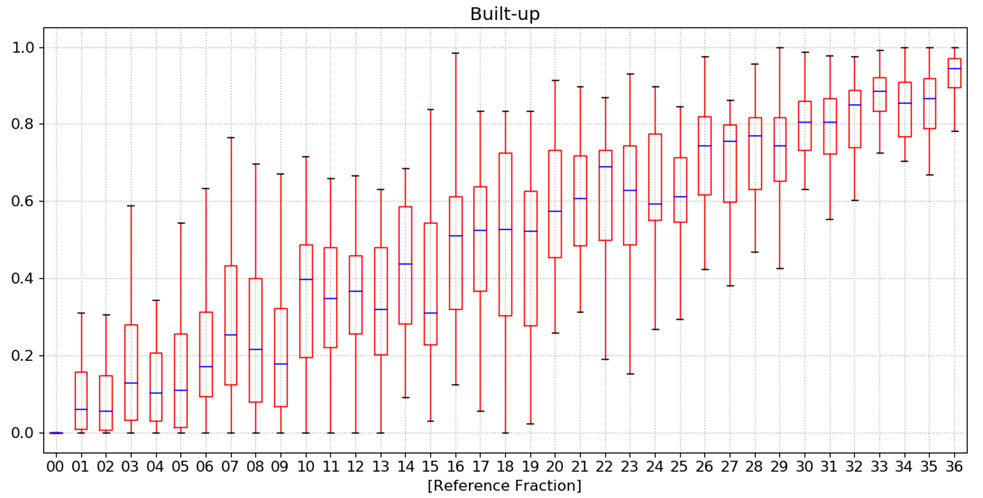}}
\subfigure[]{\includegraphics[width=1.0\textwidth]{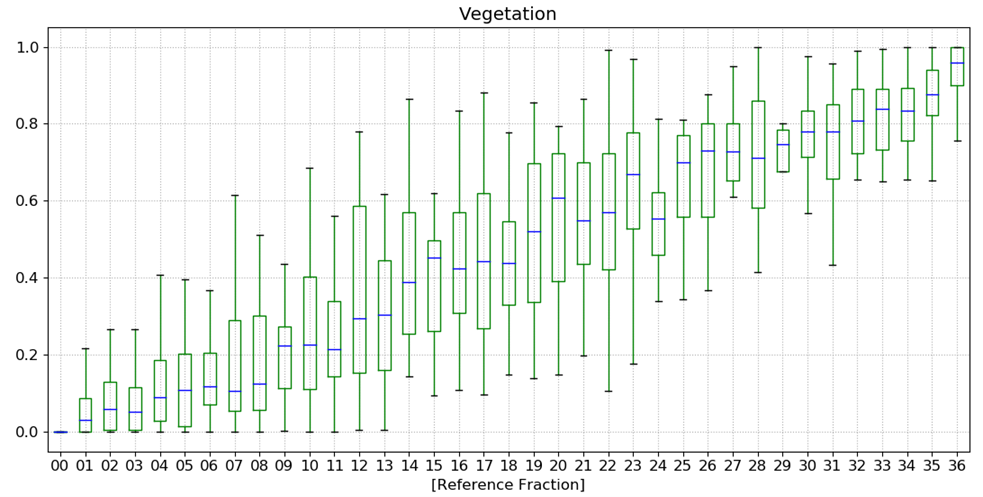}}
\caption{Box plots illustrating the spread of (a) built-up and (b) vegetation predictions corresponding to each of the reference fractions.}
\label{fig:fig9}
\end{figure*}

In this study, reference land cover fractions are computed as the ratio of number of hard classified LISS-IV cells of a particular land cover and total number of LISS-IV cells (36), corresponding to each 30m cell of the Landsat-5 TM data. Hence, the reference fractions can only have discrete floating-point values between 0 and 1 (0/36, 1/36, 2/36, …, 36/36). We can denote these fractions using two-digit numerals, i.e., (1/36) as ‘01’, (2/36) as ‘02’ and so on. Box plots in Figure \ref{fig:fig9} shows the distribution of predicted land cover fractions corresponding to each of the reference fractions. For example, box plot ‘00’ for built-up shows the spread of predicted built-up fractions when the reference fraction is 0.00. Similarly, box plot ‘36’ for built-up shows the spread of predicted built-up fractions when the reference fraction is 1.00. We have identified this as a better way of describing the characteristics of the relationship rather than using scatter plots as plotting many data points directly on a scatter plot may lead to overlapping points, which can be visually unpleasant, or even obscure the data.

Figure \ref{fig:figA1} shows two pairs of box plots that compare the reference and predicted fractions of built-up and vegetation as whole. The true medians do not differ much in both the plots indicate lesser variability between the reference and predicted values. The minimum and first quartile (Q1) line falls on the zero-point due to approximately 25\% of the area have zero built-up and vegetation fractions. For the built-up prediction, the third quartile (Q3) line falls slightly below as compared to the reference fraction which shows an under-estimation at higher values. In contrast, a minor over-estimation is observed in case of vegetation. It should be noted that the maximum values are clipped at ‘1’ in both the plots.

Figure \ref{fig:figA2}(a) and Figure \ref{fig:figA2}(b) shows the bias histograms for built-up and vegetation, respectively. For built-up, 78\% of the total predictions are within +/- 15\% of the true value and 91\% of the total predictions are within +/- 25\% of the true value. Similarly, for vegetation, 81\% of the total predictions are within +/- 15\% of the true value and 93\% of the total predictions are within +/-25\% of the true value. The maximum prediction error is exceeding ±0.50 in less than 1\% of samples for both built-up and vegetation.

\subsubsection{Receiver Operating Characteristic (ROC) curve analysis}
\begin{figure*}[ht]
\centering
\subfigure[]{\includegraphics[scale=0.95]{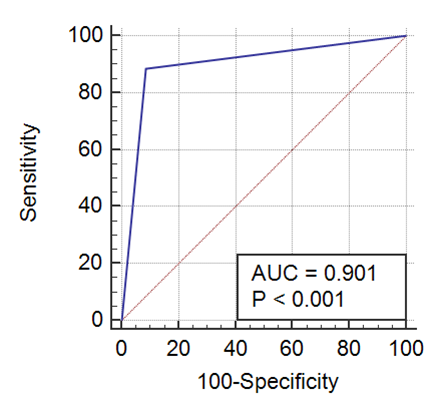}}
\subfigure[]{\includegraphics[scale=0.95]{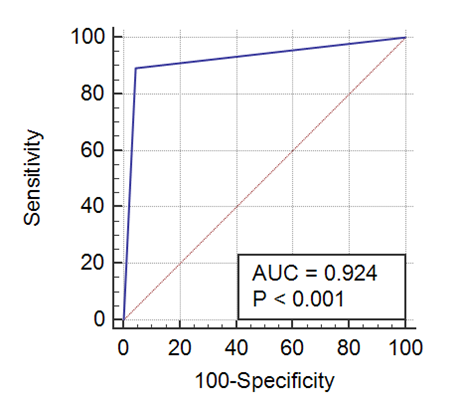}}
\caption{ROC curves at 50\% threshold level for (a) built-up and (b) vegetation.}
\label{fig:fig12}
\end{figure*}

We have performed the ROC analysis on the built-up and vegetation data to illustrate the capability of our model in doing a binary classification. For this, we set a 50\% threshold which means Landsat pixels with a predicted built-up fraction of 0.50 or above will be considered as built-up and rest will be non-built-up. Similarly, all vegetation fractions greater than 0.50 will be considered as vegetation and rest will be not vegetation. The results of ROC analysis are given in Table \ref{tableA} and the corresponding ROC curves are shown in Figure \ref{fig:fig12}.

\subsection{Generalizability of the CNN model}\label{sec4.4}
\begin{figure*}[ht!]
\centering
\includegraphics[width=15cm]{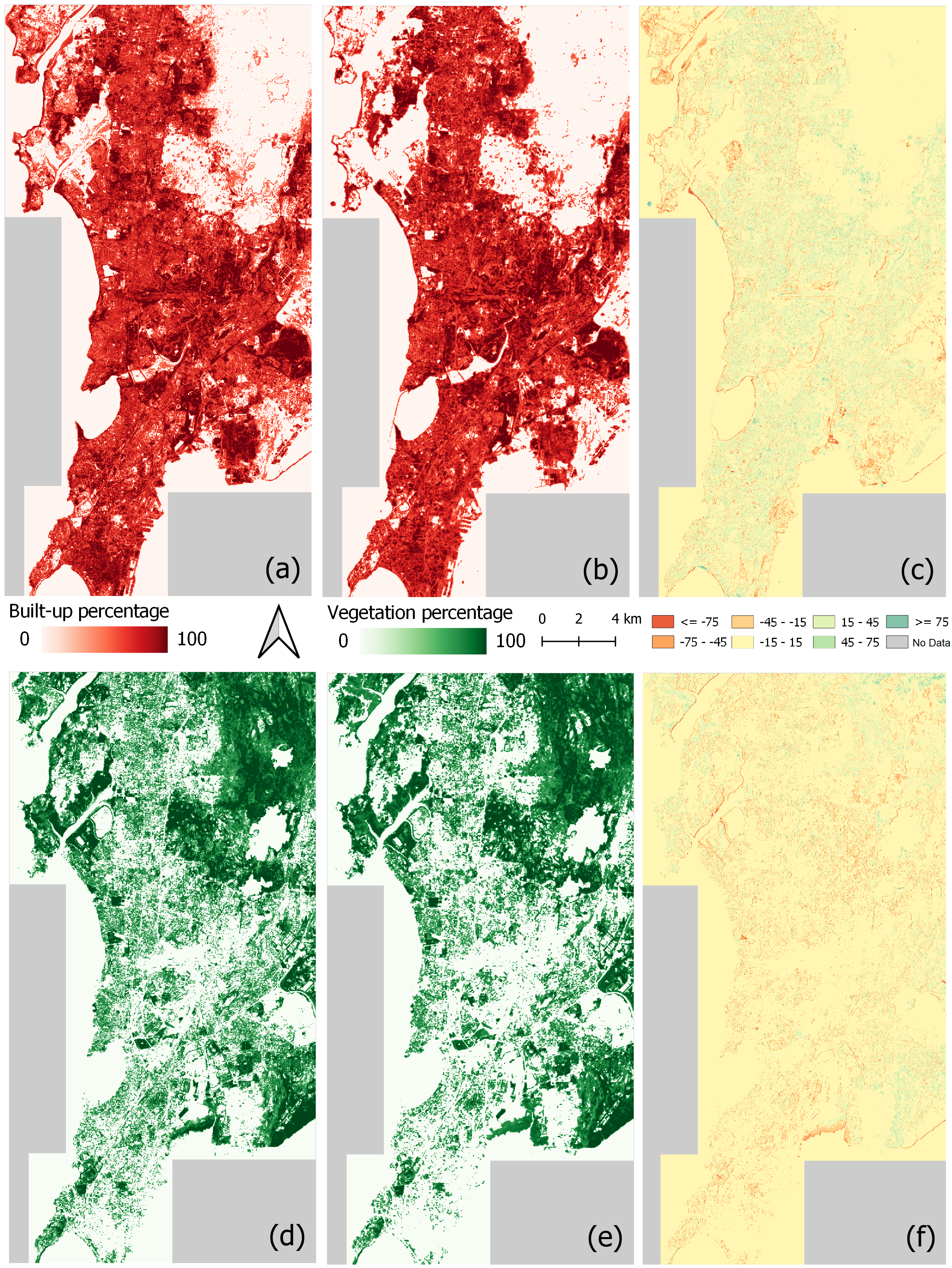}
\caption{30m resolution fractional maps of built-up and vegetation for Mumbai (2009); (a) reference built-up fraction map; (b) predicted built-up fraction map; (c) prediction error map for built-up; (d) reference vegetation fraction map; (e) predicted vegetation fraction map; (f) prediction error map for vegetation.}
\label{fig:fig13}
\end{figure*}

We used another image pair (Landsat-5 TM and LISS-IV) for Mumbai city (acquired in February 2009) for demonstrating the generalizability of our CNN model. The size of original Landsat-5 TM image was 558x1084. Since Mumbai is a coastal city and a large part of the scene is covered by open water, we have masked out most areas with water while making predictions. These masked out areas are indicated as ‘No Data’ in Figure \ref{fig:fig13}. Thus, we made predictions for only 490164 samples using the previously trained CNN model. While Figure \ref{fig:fig13} visually compares the reference and predicted fractional maps of built-up and vegetation, quantitative results are given in Table \ref{table5}. Further, the bias histograms of built-up and vegetation separately are shown in Figure \ref{fig:figA3}. For built-up, 72\% of the total predictions are within +/- 15\% of the true value and 83\% of the total predictions are within +/- 25\% of the true value. Similarly, for vegetation, 80\% of the total predictions are within +/- 15\% of the true value and 89\% of the total predictions are within +/-25\% of the true value. The bias histograms for Mumbai are therefore comparable to those obtained for Bengaluru.  As mentioned above, a large part of the Landsat-5 TM and LISS-IV scene for Mumbai is open water with zero built-up or vegetation. The model is highly accurate for these cells and hence there is a large spike in the 0.0\% to 0.05\% bin of the bias histograms.

\begin{table}[]
\centering
\caption{Performance of the CNN model on Mumbai (2009) data.}
\label{table5}
\begin{tabular}{|l|l|c|c|c|c|}
\hline
\multicolumn{1}{|c|}{Pixel   size}                                                    & Landcover type & MAE (\%) & RMSE (\%) & R2    & NSI  \\ \hline
\multirow{2}{*}{\begin{tabular}[c]{@{}l@{}}30 m × 30 m\\ (single pixel)\end{tabular}} & Built-up       & 11.3     & 19.4      & 0.78* & 0.75 \\ \cline{2-6} 
                                                                                      & Vegetation     & 7.9      & 15.6      & 0.79* & 0.78 \\ \hline
\multirow{2}{*}{\begin{tabular}[c]{@{}l@{}}90 m × 90 m\\ (3×3 pixels)\end{tabular}}   & Built-up       & 9.9      & 14.9      & 0.90* & 0.82 \\ \cline{2-6} 
                                                                                      & Vegetation     & 7.9      & 12.1      & 0.90* & 0.85 \\ \hline
\end{tabular}

\begin{tablenotes}
\centering
      \small
      \item *Statistically significant relationship $(p < 0.01)$
    \end{tablenotes}

\end{table}

\subsection{Performance Comparison with the Random Forest Regression Model}\label{sec4.5}
The performance of the CNN model is experimentally compared with a Random Forest (RF) regression model which is an ensemble learning method operates by constructing several decision trees. The RF method uses bagging or bootstrap aggregation to yield an ensemble of regression trees. The potential benefit is that it is insensitive to overfitting. For our experiment, 100 trees were constructed. Out of bag (OOB) score was 0.8354 while considering 198 features and 12000 original training samples. These 198 features include the 7x7 neighborhood of the ith pixel in first 4 spectral bands (B1-B4) and the ith pixel value in EBBI image and band 7. The results with different number of training samples were given in Table \ref{table6}. Furthermore, we have experimented with different combination of number of trees and features to ensure the best results.

The results demonstrate the superior performance of the CNN based sub-pixel land cover classification system over RF regression model. From Table \ref{table6}, it can be observed that, for Bengaluru data, the best percentage RMSE values of 16.4 (Built-up) and 12.7 (Vegetation) are obtained when the number of training samples are increased to 50000. The corresponding values for the proposed CNN model were 13.9 and 12.8 (refer Table \ref{table2}) obtained using only 12000 training samples. This implies that RF model achieved a marginal improvement in performance only for the vegetation class with supply of 38000 additional training samples. RF model produced percentage RMSE values of 17.8 and 13.9, respectively for the built-up and vegetation class when using only 12000 training samples. However, RF model is not able to compete with the CNN based model in terms of generalizability which is evident from the results on 490164 test samples of Mumbai 2009 data. Percentage RMSE values of 25.4 and 21.9, respectively are obtained with the best RF model which are considerably higher than the RMSE of 19.4 and 15.6 (refer Table \ref{table5}) obtained using the CNN model.

\vspace*{-.5cm}

\begin{table}[b!]
\caption{Performance of the RF regression model with different sets of training samples.}
\label{table6}
\resizebox{\textwidth}{!}{%
\begin{tabular}{|cccccc|cccc|}
\hline
\multicolumn{6}{|c|}{\begin{tabular}[c]{@{}c@{}}Bengaluru   (2011)\\ (2250 test samples)\end{tabular}}                                                                                                                                                                                                                                                 & \multicolumn{4}{c|}{\begin{tabular}[c]{@{}c@{}}Mumbai (2009)\\ (490164 test samples)\end{tabular}}                                                                               \\ \hline
\multicolumn{1}{|c|}{\begin{tabular}[c]{@{}c@{}}\#Training\\ samples\end{tabular}} & \multicolumn{1}{c|}{\begin{tabular}[c]{@{}c@{}}Land cover\\ type\end{tabular}} & \multicolumn{1}{c|}{\begin{tabular}[c]{@{}c@{}}MAE\\ (\%)\end{tabular}} & \multicolumn{1}{c|}{\begin{tabular}[c]{@{}c@{}}RMSE\\ (\%)\end{tabular}} & \multicolumn{1}{c|}{NSI}  & R2    & \multicolumn{1}{c|}{\begin{tabular}[c]{@{}c@{}}MAE\\ (\%)\end{tabular}} & \multicolumn{1}{c|}{\begin{tabular}[c]{@{}c@{}}RMSE\\ (\%)\end{tabular}} & \multicolumn{1}{c|}{NSI}  & R2    \\ \hline
\multicolumn{1}{|c|}{\multirow{2}{*}{12000}}                                       & \multicolumn{1}{c|}{Built-up}                                                  & \multicolumn{1}{c|}{11.5}                                               & \multicolumn{1}{c|}{17.8}                                                & \multicolumn{1}{c|}{0.81} & 0.81* & \multicolumn{1}{c|}{20.3}                                               & \multicolumn{1}{c|}{26.9}                                                & \multicolumn{1}{c|}{0.50} & 0.70* \\ \cline{2-10} 
\multicolumn{1}{|c|}{}                                                             & \multicolumn{1}{c|}{Vegetation}                                                & \multicolumn{1}{c|}{9.1}                                                & \multicolumn{1}{c|}{13.9}                                                & \multicolumn{1}{c|}{0.85} & 0.85* & \multicolumn{1}{c|}{15.1}                                               & \multicolumn{1}{c|}{22.4}                                                & \multicolumn{1}{c|}{0.61} & 0.65* \\ \hline
\multicolumn{1}{|c|}{\multirow{2}{*}{15000}}                                       & \multicolumn{1}{c|}{Built-up}                                                  & \multicolumn{1}{c|}{11.2}                                               & \multicolumn{1}{c|}{16.9}                                                & \multicolumn{1}{c|}{0.83} & 0.84* & \multicolumn{1}{c|}{20.7}                                               & \multicolumn{1}{c|}{27.1}                                                & \multicolumn{1}{c|}{0.49} & 0.70* \\ \cline{2-10} 
\multicolumn{1}{|c|}{}                                                             & \multicolumn{1}{c|}{Vegetation}                                                & \multicolumn{1}{c|}{9.3}                                                & \multicolumn{1}{c|}{14.2}                                                & \multicolumn{1}{c|}{0.84} & 0.84* & \multicolumn{1}{c|}{14.7}                                               & \multicolumn{1}{c|}{22.1}                                                & \multicolumn{1}{c|}{0.62} & 0.67* \\ \hline
\multicolumn{1}{|c|}{\multirow{2}{*}{20000}}                                       & \multicolumn{1}{c|}{Built-up}                                                  & \multicolumn{1}{c|}{11.4}                                               & \multicolumn{1}{c|}{17.1}                                                & \multicolumn{1}{c|}{0.83} & 0.83* & \multicolumn{1}{c|}{19.1}                                               & \multicolumn{1}{c|}{25.7}                                                & \multicolumn{1}{c|}{0.54} & 0.72* \\ \cline{2-10} 
\multicolumn{1}{|c|}{}                                                             & \multicolumn{1}{c|}{Vegetation}                                                & \multicolumn{1}{c|}{9.1}                                                & \multicolumn{1}{c|}{14.0}                                                & \multicolumn{1}{c|}{0.84} & 0.84* & \multicolumn{1}{c|}{14.7}                                               & \multicolumn{1}{c|}{22.2}                                                & \multicolumn{1}{c|}{0.62} & 0.66* \\ \hline
\multicolumn{1}{|c|}{\multirow{2}{*}{25000}}                                       & \multicolumn{1}{c|}{Built-up}                                                  & \multicolumn{1}{c|}{11.9}                                               & \multicolumn{1}{c|}{18.2}                                                & \multicolumn{1}{c|}{0.80} & 0.81* & \multicolumn{1}{c|}{20.2}                                               & \multicolumn{1}{c|}{26.7}                                                & \multicolumn{1}{c|}{0.50} & 0.70* \\ \cline{2-10} 
\multicolumn{1}{|c|}{}                                                             & \multicolumn{1}{c|}{Vegetation}                                                & \multicolumn{1}{c|}{9.0}                                                & \multicolumn{1}{c|}{13.6}                                                & \multicolumn{1}{c|}{0.85} & 0.85* & \multicolumn{1}{c|}{14.4}                                               & \multicolumn{1}{c|}{22.0}                                                & \multicolumn{1}{c|}{0.63} & 0.68* \\ \hline
\multicolumn{1}{|c|}{\multirow{2}{*}{50000}}                                       & \multicolumn{1}{c|}{Built-up}                                                  & \multicolumn{1}{c|}{10.5}                                               & \multicolumn{1}{c|}{16.4}                                                & \multicolumn{1}{c|}{0.84} & 0.84* & \multicolumn{1}{c|}{19.3}                                               & \multicolumn{1}{c|}{25.4}                                                & \multicolumn{1}{c|}{0.55} & 0.71* \\ \cline{2-10} 
\multicolumn{1}{|c|}{}                                                             & \multicolumn{1}{c|}{Vegetation}                                                & \multicolumn{1}{c|}{8.3}                                                & \multicolumn{1}{c|}{12.7}                                                & \multicolumn{1}{c|}{0.87} & 0.87* & \multicolumn{1}{c|}{14.6}                                               & \multicolumn{1}{c|}{21.9}                                                & \multicolumn{1}{c|}{0.63} & 0.67* \\ \hline
\end{tabular}%
}
\begin{tablenotes}
\centering
      \small
      \item *Statistically significant relationship $(p < 0.01)$
\end{tablenotes}

\end{table}

\vspace*{3cm}
\section{Discussion}\label{sec5}
\vspace*{-.1cm}
This study is substantially different from its predecessors \shortcite{deng2020continuous,li2020mapping,patidar2018multi,deng2017subpixel,maclachlan2017subpixel} in that it is conducted over a much larger spatial extent and does not rely on any temporal information. Moreover, we have tested and validated our model on Landsat images acquired on different time periods (2011 and 2009) and different geographical areas (Bengaluru and Mumbai). Predictions are made on 490164 samples (30mx30m Landsat cells) obtained from a large spatial extent of Mumbai city where a lot of variations in the landscape pattern and pixel heterogeneity can be observed. Further, to prove the superior performance of the proposed CNN model, we experimentally compared it against the simple linear regression model and the advanced RF regression model. We believe that training the model with randomly collected data from a large spatial extent helped us to achieve greater generalizability though more research is required to prove this.

\vspace*{-.1cm}
\subsection{Comparison with Recent Models }\label{sec5.1}
Several methods have been proposed in the literature to unmix the pixels of a coarse resolution image. All these methods have used some high-resolution reference image for training their model and evaluating the predicted fractions. Hence, depending on the quality (improved resolution) of the reference image, the prediction accuracy may differ. Further, there exist multiple factors which can impact the accuracy of estimated fractional landcover such as atmospheric condition during image acquisition, environmental and urban settings, seasonal variation, and inclusion of temporal features \shortcite{deng2017subpixel}. Table \ref{table7} summarizes the performance of recently proposed sub-pixel classification models and compare it with the proposed method. It should be noted that we have included only those studies in Table \ref{table7} which compute impervious surface fraction (or similar class such as urban or built-up) and/or vegetation fraction from Landsat images.
	
Spectral mixture analysis (SMA) was employed in many studies for mapping the abundances of land covers. \shortciteA{shih2020estimating}, to map impervious change from 1990 to 2015, used the SMA method to unmix the Landsat time series data including different sensors. However, a single dated imagery has been used for the endmember selection. \%RMSE of this study ranged between 15.8 to 28.9 for impervious surface and 19.0 to 34.0 for vegetation. The authors pointed out multiple reasons for this broad error range including vegetation seasonality variation, high-resolution image registration errors, insufficient endmembers and cloud covers in the scene. \shortciteA{li2020mapping} integrated spectral indices with a liner constrained SMA to map impervious surfaces distributions from a Landsat-5 TM imagery. The spectral indices were used for endmember selection and extraction. The authors claim that this integration works better than the conventional SMA. \shortciteA{patidar2018multi} integrated two linear SMA based techniques, viz. Normalized Spectral Mixture Analysis (NSMA) and Multiple Endmember Spectral Mixture Analysis (MESMA) to estimate land cover fractions from a Landsat image. However, the authors have later combined the outputs of unmixing by multiple methods including support vector regression (SVR) and multi-layer perceptron (MLP) using a Bayesian model ensemble technique to achieve better results. \%RMSE values of 10.2 for impervious surface (built-up) and 11.81 for vegetation are reported using test pixel size of 90mx90m. For the same test pixel size, our model has a slightly higher \%RMSE value for impervious surface (10.5) as shown in Table \ref{table7}, however, it performs significantly better in the case of vegetation (7.2). The same authors proposed an unmixing method \shortcite{patidar2020rule} to derive sub-pixel impervious surface fraction at annual scale using Landsat TM scenes by integrating temporal contextual information into MESMA method. The performance of this method was much better compared to the former method. However, this method used some temporal features extracted from a time series data which can hardly be derived from a single-date image or even seasonal combinations. Hence, a direct comparison between this method and the proposed one is difficult as we are not using any time series data or temporal filtering.

While SMA is still regarded as one powerful approach for the estimation of impervious surfaces, it has some limitations. It demands highly accurate endmember selection and for this we need a separate and robust algorithm. For example, \shortciteA{shih2020estimating} used visual inspection of endmember spectra and K-means clustering for the endmember selection. Here, the parameter K needs to be estimated for each land cover class and this process was not straightforward. Due to spatial variations of physical structures and complex geographic conditions, endmember spectra can vary and hence, a fixed endmember spectrum may not be always enough. Further, this makes it difficult for the same model to be used for a different locality. To partially overcome these limitations, \shortciteA{patidar2020rule} employed a temporal filtering method under the assumption that urbanization is an irreversible process and impervious surface cannot change to vegetation or soil. However, this may not be valid in instances where large industrial or other buildings get redeveloped into parks or residences.

Previous studies \shortcite{yuan2008comparison,deng2013use} have shown that regression tree and RF models to yield better results than spectral mixture analysis for impervious surface estimation when enough reference data is available. \shortciteA{deng2020continuous} proposed a continuous subpixel monitoring scheme using Landsat time series data to quantify the impervious surface change. Though this method used an RF regressor to estimate the fractional land cover, the input variables of the RF model were coefficients and errors derived from the time series model. The same authors \shortcite{deng2017subpixel} previously used an RF model to extract impervious surface fraction from a single-date Landsat image which reports \%RMSE values in the range 8.2 to 11.9 with the test pixel size of 90mx90m. Hence, only the latter approach is included in Table \ref{table7} for comparison due to previously stated reasons regarding temporal features.

 \shortciteA{maclachlan2017subpixel} performed both hard and soft classification of Landsat-5 TM image using import vector machine (IVM). IVM produces probability distributions of land cover classes instead of deterministic fractions. The classification accuracy is evaluated using Google Earth and aerial images. For the hard classification, \shortciteA{maclachlan2017subpixel} achieved 84\% accuracy with a kappa coefficient of 0.78, while the proposed CNN model scores kappa coefficient values greater than 0.80 (refer Table \ref{tableA}). However, for the soft classification, \shortciteA{maclachlan2017subpixel} reported higher error values for the urban area due to overestimation. \shortciteA{deng2019examining} tried to unmix a Landsat-5 TM image using a deep belief network (DBN). Like IVM, the outputs of DBN are probability distributions of land cover classes. Since it was difficult to use mixed pixels to train DBN, the samples were collected with the help of a 1m resolution AISA hyper-spectral image.  The AISA image was earlier resampled to match with the wavelengths of Landsat image. The best \%RMSE for DBN was 7.7 and the authors have found DBN providing comparable results with RF, MESMA, and support vector machine (SVM).

From the discussion above and Table \ref{table7}, it can conclude that the performance of the proposed CNN model is quite satisfactory and is better or equal to the recently proposed sub-pixel classification methods. However, there is still a potential for improvement in this method by improving the quality of the reference data. While the above discussed models used reference images of resolution less than or equal to 1m (refer Table \ref{table7}), we were relying on 5m resolution LISS IV imagery which was procured at a relatively low cost. Another strength of our model is that we are not integrating any temporal information into the classification model as there are both pros and cons identified in this technique. Though it results in an improved performance of the model in estimating the landcover fractions from a time series data, the same model can hardly be used for making predictions from a single dated image for a different location. Since, the proposed method does not rely on any temporal information, it can be used for making predictions on a single Landsat-5 TM image acquired on any date.

\subsection{Generation of Time Series of Landcover Data}\label{sec5.2}
By predicting land cover fractions from a Landsat image for a different city (Mumbai) acquired on a distinct year (2009), we demonstrated that our model is suitable for assessing the land cover changes for the years for which Landsat-5 TM data is available. This is especially significant since availability of high-resolution imagery is limited for the years prior to 2000. Using our trained model, it is possible to predict the land cover proportion of previous years at sub-pixel resolution although Landsat-5 has a coarse spatial resolution.  Since Landsat-5 TM was operating from 1985 till 2011, it will be possible to generate a time series of sub-pixel land cover maps for this period using our model. This will help analyze how areas with varying built-up densities within a city have grown over time thereby enabling a neighborhood level analysis of the nature and extent of urban expansion, and spatial-temporal patterns of urban land cover transitions and resulting impacts on communities, ecosystems, and microclimate \shortcite{schug2018mapping,sultana2020assessment,ridd1995exploring}. The sub-pixel level built-up information have further applications such as downscaling the spatial distribution of population and income levels to a finer scale \shortcite{azar2013generation}.

\subsection{Limitations of this Study}\label{sec5.3}
The accuracy assessment of the sub-pixel classification results was done with respect to the reference data. However, the accuracy of the reference data itself is not 100\% guaranteed since it is produced from the manual classification of a high-resolution image by an expert. The absence of a perfect reference data to train or validate the model is a general problem in remote sensing which arises from the impracticality of the ground truth reference data collection by extensive field surveys. \shortciteA{williams2017validation} suggest that all forms of reference data require validation prior to use in assessing the performance of classification and/or unmixing algorithms. Here, we discuss two solutions to this problem. One is to collect the reference data using a VHR imagery with spatial resolution less than 1m. This can substantially avoid the mixed pixels in the image and thereby improve the classification accuracy. However, the availability of VHR imagery may be a constraint which is again limited by high data costs. Another solution is to classify the available high-resolution image visually (pixel by pixel) by multiple experts and prepare the reference data based on the mutual agreement between them. Even though this process appears slow and tedious, it can generate reference data with near 100\% accuracy.

Another limitation of this study is that one must match the geometry of the two images to make them perfectly overlap with each other at pixel accuracy such that image registration errors are minimized. This can create marginal errors if the spatial resolution values are not perfectly divisible. Here, we have resampled the LISS-IV imagery to 5m resolution, such that the spatial resolution of the Landsat image (30 m) is perfectly divisible by the resampled LISS-IV image resolution. But such resampling can also be a source of errors.

\section{Conclusions}\label{sec6}
A CNN model is designed for the sub-pixel estimation of built-up and vegetation in Landsat-5 TM data by using a hard classified LISS-IV image for training. We trained the model using the satellite data of Bengaluru (2011) and demonstrated the generalizability of the method using the data of Mumbai (2009). Thus, we have succeeded in simulating the sensor behavior of Landsat-5 TM and demonstrated that we can go back in time to generate the sub-pixel maps for other years when Landsat-5 TM was active. The proposed work will contribute to monitoring land cover change at finer scale and capturing the heterogeneity in urban areas better. Our future work will focus on generating the sub-pixel level land cover maps for different years from 1985-2011 for temporal analysis of the urban landscape.

\begin{landscape}
\begin{table}
\caption{Performance comparison of recent sub-pixel classification models.}
\label{table7}
\resizebox{\columnwidth}{!}{%
\begin{tabular}{|l|l|l|l|cc|c|c|cc|}
\hline
\multicolumn{1}{|c|}{\multirow{2}{*}{\#}} & \multicolumn{1}{c|}{\multirow{2}{*}{Year}} & \multicolumn{1}{c|}{\multirow{2}{*}{Author}} & \multicolumn{1}{c|}{\multirow{2}{*}{Approach}} & \multicolumn{2}{c|}{Satellite Datasets}                                                                            & \multirow{2}{*}{Pixel size}                   & \multirow{2}{*}{\begin{tabular}[c]{@{}c@{}}Land cover\\ type\end{tabular}} & \multicolumn{2}{c|}{Performance}               \\ \cline{5-6} \cline{9-10} 
\multicolumn{1}{|c|}{}                    & \multicolumn{1}{c|}{}                      & \multicolumn{1}{c|}{}                        & \multicolumn{1}{c|}{}                          & \multicolumn{1}{c|}{Coarse Resolution}               & High Resolution                                             &                                               &                                                                            & \multicolumn{1}{c|}{\%RMSE}      & \%MAE       \\ \hline
\multirow{2}{*}{1}                        & \multirow{2}{*}{2020}                      & \multirow{2}{*}{Shih   et al.}               & \multirow{2}{*}{SMA}                           & \multicolumn{1}{l|}{\multirow{2}{*}{Landsat series}} & \multicolumn{1}{l|}{\multirow{2}{*}{GE and Aerial imagery}} & \multicolumn{1}{l|}{\multirow{2}{*}{90mx90m}} & IS                                                                         & \multicolumn{1}{c|}{15.8 – 28.9} & 11.3 – 21.1 \\ \cline{8-10} 
                                          &                                            &                                              &                                                & \multicolumn{1}{l|}{}                                & \multicolumn{1}{l|}{}                                       & \multicolumn{1}{l|}{}                         & V                                                                          & \multicolumn{1}{c|}{19.0 – 34.0} & 13.4 – 29.3 \\ \hline
2                                         & 2019                                       & Li                                           & Linear constrained SMA                         & \multicolumn{1}{c|}{Landsat-5 TM}                    & GE and Aerial imagery                                       & \multicolumn{1}{l|}{30mx30m}                  & IS                                                                         & \multicolumn{1}{c|}{-}           & 10.1 – 14.2 \\ \hline
3                                         & 2019                                       & Deng   et al.                                & DBN                                            & \multicolumn{1}{c|}{Landsat-5 TM}                    & GE and AISA(1m)                                             & \multicolumn{1}{l|}{90mx90m}                  & IS                                                                         & \multicolumn{1}{c|}{7.7}         & 6.0         \\ \hline
\multirow{2}{*}{4}                        & \multirow{2}{*}{2018}                      & \multirow{2}{*}{Patidar   and Keshari}       & \multirow{2}{*}{SVR, MLP and SMA}              & \multicolumn{1}{c|}{\multirow{2}{*}{Landsat ETM+}}   & \multirow{2}{*}{OrbitView-3 (1m)}                           & \multicolumn{1}{l|}{\multirow{2}{*}{90mx90m}} & IS                                                                         & \multicolumn{1}{c|}{10.2}        & -           \\ \cline{8-10} 
                                          &                                            &                                              &                                                & \multicolumn{1}{c|}{}                                &                                                             & \multicolumn{1}{l|}{}                         & V                                                                          & \multicolumn{1}{c|}{11.8}        & -           \\ \hline
5                                         & 2017                                       & Deng   et al.                                & RF                                             & \multicolumn{1}{c|}{Landsat ETM+}                    & Aerial imagery                                              & 90mx90m                                       & IS                                                                         & \multicolumn{1}{c|}{8.2 – 11.9}  & 4.6 – 7.8   \\ \hline
\multirow{4}{*}{6}                        & \multirow{4}{*}{2017}                      & \multirow{4}{*}{MacLachlan   et al.}         & \multirow{4}{*}{IVM}                           & \multicolumn{1}{c|}{\multirow{4}{*}{Landsat-5 TM}}   & \multirow{4}{*}{GE and aerial imagery}                      & \multirow{2}{*}{30mx30m}                      & IS                                                                         & \multicolumn{1}{c|}{31.8}        & -           \\ \cline{8-10} 
                                          &                                            &                                              &                                                & \multicolumn{1}{c|}{}                                &                                                             &                                               & V                                                                          & \multicolumn{1}{c|}{12.1}        & -           \\ \cline{7-10} 
                                          &                                            &                                              &                                                & \multicolumn{1}{c|}{}                                &                                                             & \multirow{2}{*}{90mx90m}                      & IS                                                                         & \multicolumn{1}{c|}{22.0}        & -           \\ \cline{8-10} 
                                          &                                            &                                              &                                                & \multicolumn{1}{c|}{}                                &                                                             &                                               & V                                                                          & \multicolumn{1}{c|}{7.9}         & -           \\ \hline
\multirow{4}{*}{7}                        & \multirow{4}{*}{2021}                      & \multirow{4}{*}{Proposed   Method}           & \multirow{4}{*}{CNN}                           & \multicolumn{1}{c|}{\multirow{4}{*}{Landsat-5   TM}} & \multirow{4}{*}{Resourcesat-1   LISS-IV}                    & \multirow{2}{*}{30mx30m}                      & IS                                                                         & \multicolumn{1}{c|}{13.9}        & 9.1         \\ \cline{8-10} 
                                          &                                            &                                              &                                                & \multicolumn{1}{c|}{}                                &                                                             &                                               & V                                                                          & \multicolumn{1}{c|}{12.8}        & 7.2         \\ \cline{7-10} 
                                          &                                            &                                              &                                                & \multicolumn{1}{c|}{}                                &                                                             & \multirow{2}{*}{90mx90m}                      & B                                                                          & \multicolumn{1}{c|}{10.5}        & 6.0         \\ \cline{8-10} 
                                          &                                            &                                              &                                                & \multicolumn{1}{c|}{}                                &                                                             &                                               & V                                                                          & \multicolumn{1}{c|}{7.2}         & 4.5         \\ \hline
\end{tabular}%
}
\begin{tablenotes}
      \small
      \item CNN: Convolutional Neural Network; DBN: Deep Belief Network;  IVM: Import Vector Machine;  MLP: Multi-Layer Perceptron;  RF: Random Forest;
      \item SMA: Spectral Mixture Analysis;  SVR: Support Vector Regression;  IS: Impervious Surface (or Urban or Built-up);	B:Built-up; V: Vegetation;	GE: Google Earth.
    \end{tablenotes}

\end{table}
\end{landscape}

\newpage
\begin{appendices}
\setcounter{figure}{0}
\renewcommand{\thefigure}{A\arabic{figure}}

\section{Figures}\label{appendixa}

\begin{figure*}[ht!]
\centering
\includegraphics[width=15cm]{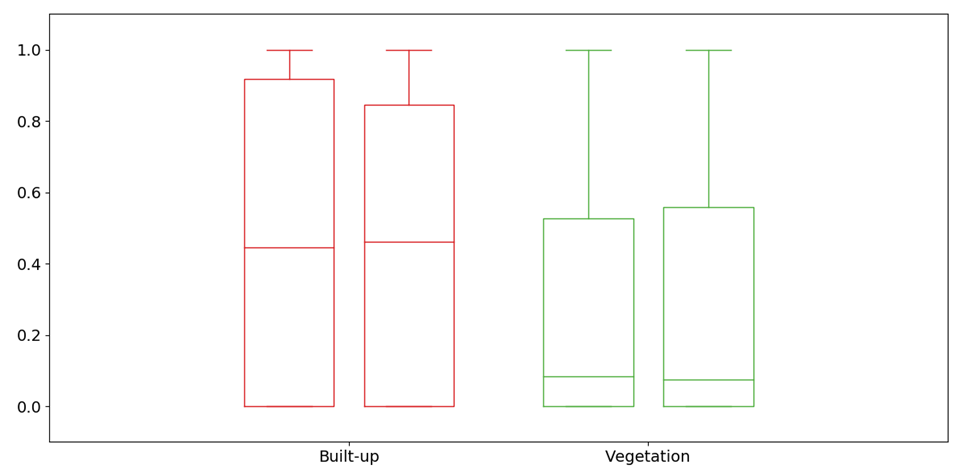}
\caption{Box plots comparing the reference (left) and predicted (right) fractions for built-up and vegetation.}
\label{fig:figA1}
\end{figure*}

\pagebreak 

\begin{figure*}[ht!]
\centering
\subfigure[]{\includegraphics[width=15cm]{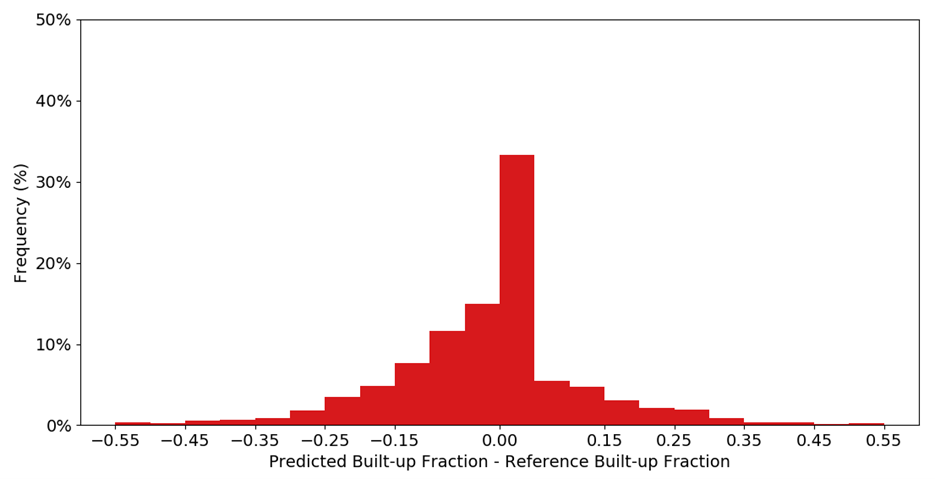}}
\subfigure[]{\includegraphics[width=15cm]{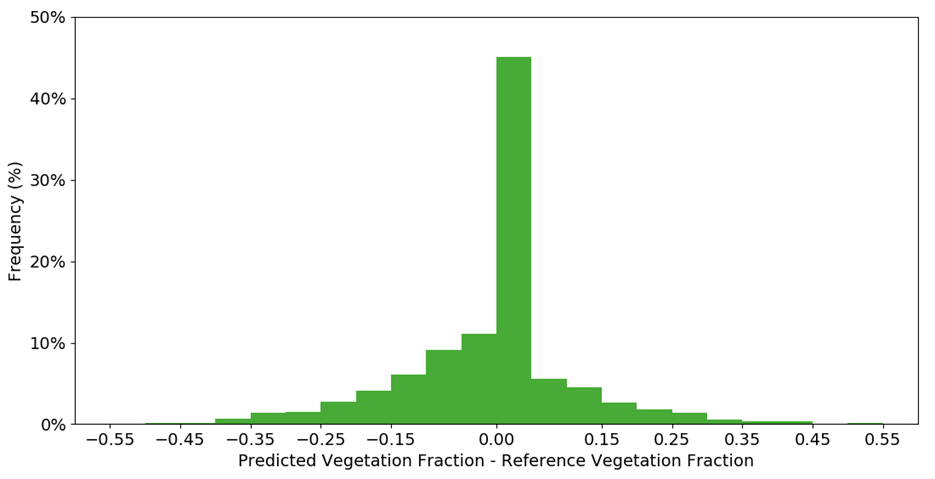}}
\caption{Bias histograms for Bengaluru showing the frequency of error values computed between each reference and predicted fraction samples for (a) built-up and (b) vegetation.}
\label{fig:figA2}
\end{figure*}

\pagebreak 

\begin{figure*}[ht!]
\centering
\subfigure[]{\includegraphics[width=15cm]{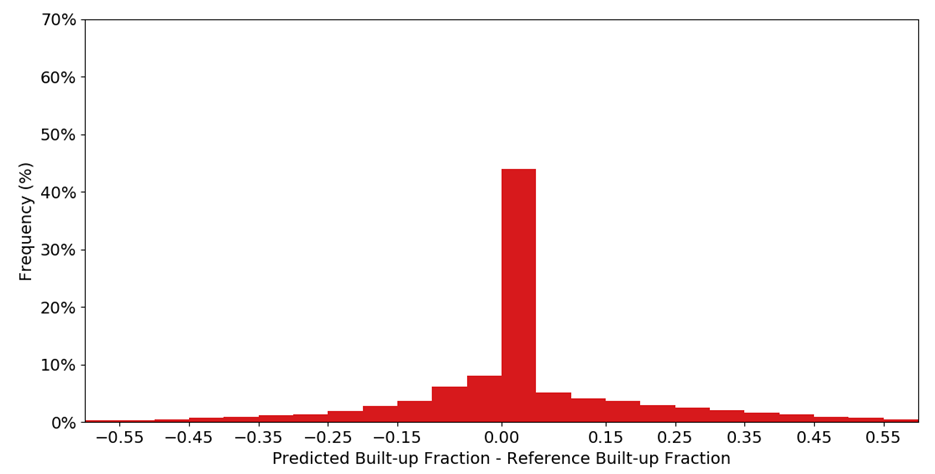}}
\subfigure[]{\includegraphics[width=15cm]{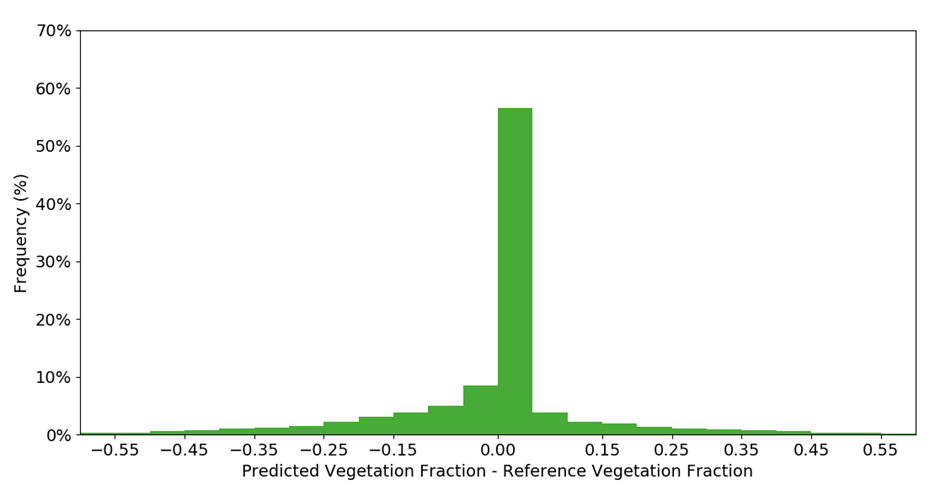}}
\caption{Bias histograms for (a) built-up and (b) vegetation for Mumbai (2009).}
\label{fig:figA3}
\end{figure*}

\pagebreak 

\setcounter{table}{0}
\renewcommand{\thetable}{B\arabic{table}}
\section{Tables}\label{appendixb}

\begin{table}[ht!]
\centering
\caption{Details of ROC analysis for built-up and vegetation.}
\label{tableA}
\begin{tabular}{|l|c|c|}
\hline
                                 & Built-up          & Vegetation        \\ \hline
Area Under the ROC Curve   (AUC) & 0.901             & 0.924             \\ \hline
Standard error                   & 0.00492           & 0.00531           \\ \hline
95\% Confidence interval         & 0.892   to 0.909  & 0.916   to 0.931  \\ \hline
Significance level P (Area=0.5)  & \textless{}0.0001 & \textless{}0.0001 \\ \hline
Sensitivity                      & 88.7              & 89.3              \\ \hline
Specificity                      & 91.5              & 95.5              \\ \hline
Cohen’s kappa coefficient        & 0.80              & 0.84              \\ \hline
\end{tabular}
\end{table}

\end{appendices}

\newpage
\bibliographystyle{apacite}
\bibliography{references}

\end{document}